%% file: main.tex
\documentclass{article}
\usepackage[preprint]{tmlr}
\usepackage[utf8]{inputenc}
\usepackage{amsmath,amssymb,amsthm}
\usepackage{microtype}
\usepackage{tabularx}
\usepackage{booktabs,graphicx}
\usepackage{xcolor}
\usepackage{hyperref}
\usepackage{enumitem}
\usepackage{placeins}
\hypersetup{hidelinks}
\ifdefined\pdfsuppressptexinfo\pdfsuppressptexinfo=-1\fi
\ifdefined\pdftrailerid\pdftrailerid{}\fi
\ifdefined\pdfinfoomitdate\pdfinfoomitdate=1\fi
\newcommand{\E}{\mathbb{E}}
\newcommand{\corr}{\mathrm{corr}}
\newtheorem{theorem}{Theorem}\newtheorem{proposition}{Proposition}
\newtheorem{corollary}{Corollary}
\newtheorem{lemma}{Lemma}
\newtheorem*{restatement}{\restatementname}
\newcommand*{\restatementname}{}
\newcommand{\tabfont}{\scriptsize}
\renewcommand{\arraystretch}{0.96}

\title{Ninety-Six Percent Correct:\\What an Evaluator's Agreement with Reality Guarantees About Decision Regret}
\author{\name Michele Viola \email m.viola.research@gmail.com}

\makeatletter\@secpenalty=0\makeatother
\begin{document}
\maketitle
\begin{abstract}
Simulators, world models, translation metrics and language-model judges stand in for expensive real
outcomes. Their agreement with those outcomes is usually reported as one correlation, and read as
license to choose: SIMPLER's repository offers its simulator, validated by a correlation and a mean
maximum rank violation (MMRV), as a source of metrics for checkpoint selection. We ask what
that number guarantees about the decisions made with the evaluator: which policy to use on each
task (routing), and which policy to deploy (deployment). A statistic's \emph{guarantee} is the largest regret, the real
success rate lost against the best choice, that any evaluator with the reported value could cause.
It is a worst-case bound, not a probability. First, with many policies a
correlation is vacuous: with $K$ policies, it guarantees anything only if $1-r^2<2.92/K$.
None of the $51$ correlations printed in eleven robot-evaluation papers guarantees routing
within five points of the best. The highest, $.995$, cannot rule out a loss of $14.4$ points.
In four larger audits, $766$ of $798$ correlations are vacuous. Second, the MMRV, which robot papers already report,
has a guarantee linear in its value. On every published table we examine, its routing guarantee computed from that table is never looser than the correlation's. As an evaluator's Gaussian errors of size $\sigma$ shrink, the
correlation's guarantee from its value falls like $\sigma$, while the expected regret and the MMRV's expected guarantee are at most of order $\sigma^2$. Third, no statistic that only compares policies within tasks can, from its value alone, certify which policy to deploy,
even at its perfect value. An evaluator can order the policies correctly on every task and still
deploy the worst one. Deployment also needs each policy's mean. We propose a validation
card that lists what to publish, and release a library that turns a reported number into the
regret it allows.
\end{abstract}

\input{intro}
\input{setting}
\input{report}
\input{exchange}
\input{card}
\input{discussion}

\input{impact}

\bibliographystyle{tmlr}
\bibliography{refs}

\appendix
\makeatletter
\let\origsection\section
\newcommand{\appsec}[1]{\origsection{#1}\suppressfloats[t]}
\newcommand{\appsecstar}[1]{\origsection*{#1}\suppressfloats[t]}
\renewcommand{\section}{\@ifstar{\appsecstar}{\appsec}}
\makeatother
\input{app_table}
\input{app_worst}

\input{app_corpus}

\input{app_audits}
\input{app_trials}
\end{document}

%% file: intro.tex
\section{Introduction}

In 1884, John Finley's experimental tornado forecasts were $96.6\%$ correct; his critics pointed out
that forecasting ``no tornado'' every time would have scored $98.2\%$ \citep{murphy1996finley}. A
pooled correlation can mislead in the same way. An evaluator that knows only which tasks are easy and
which are hard can agree closely with reality while never telling two policies apart.

Machine learning increasingly relies on evaluators that stand in for expensive real outcomes:
physics simulators and learned world models for robot trials \citep{simpler,worldgym,autoeval},
automatic metrics for professional judgements of translations \citep{freitag2024wmt24metrics},
language-model judges for human preferences \citep{dorner2025limits,landesberg2026bestofn}, and proxy
metrics in online experiments \citep{tripuraneni2024proxy}. Before an evaluator is trusted, it is
validated. $K$ candidates, which we call \emph{policies}, are run on $T$ tasks, once for real and once
in the evaluator. The agreement between the real outcomes and the evaluator's scores is then reported, usually as one correlation. A team uses the evaluator to decide which policy to use on each task (routing), which policy to deploy (deployment), or which few policies to test for real (top-$k$ screening). How much can such a decision lose, given only the
number the validation reported?

\begin{figure}[t]
\centering\includegraphics[width=\linewidth]{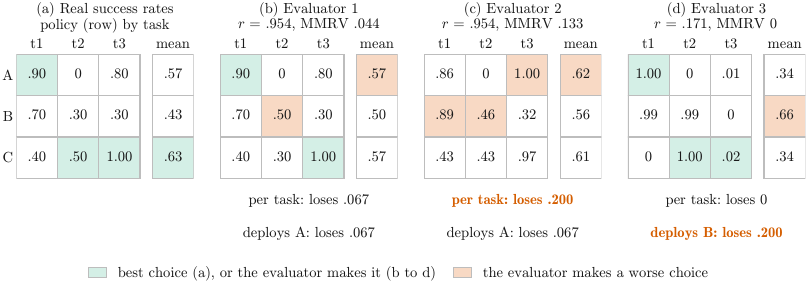}
\caption{\textbf{Each statistic is blind to one decision.} (a) Real success rates of three policies (rows) on three tasks (columns), and each policy's mean. Green marks the best policy on each task and the best mean. (b to d) Three evaluators' scores. On each task, an evaluator picks the policy it scores highest; it deploys the policy with the highest mean score. Green marks a best choice, orange a worse one. A loss is the best real rate minus the rate of the choice, averaged over tasks for routing ($.200$ is $20$ points); losses above ten points are in bold. Ties are broken against the evaluator, as for the means of A and C in (b). Evaluators 1 and 2 both have a Pearson correlation of about $.954$ with (a). Per task, the first loses $.067$ and the second $.200$, more than using the best-mean policy C everywhere ($.167$). Their MMRVs, $.044$ and $.133$, differ by the same factor of three as their losses. Evaluator 3 orders the policies on every task correctly (MMRV $0$) and loses nothing per task, yet it deploys B, the worst policy, and loses $.200$; its correlation is only $.171$. }
\label{fig:example}
\end{figure}

Figure~\ref{fig:example} shows that a correlation does not say. Evaluators 1 and 2 have the same
correlation with the real table to three digits, $.954$. Yet the first picks the best policy on two of the three
tasks and the second on none, so the second loses three times as much. The choice on each task depends only on the order of the policies within that task, not on their levels across tasks. Statistics that
compare policies only within each task fail the other way. One of them is the mean maximum rank
violation (MMRV) of SIMPLER \citep{simpler}, which robot validations report. Evaluator~3 orders the policies correctly on every task, so its MMRV is $0$. Yet it deploys the worst policy: deployment depends on levels across tasks, which such statistics never see.

These numbers are read as license to choose. SIMPLER validates its simulator by a correlation and an
MMRV, and its public repository\footnote{\url{https://github.com/simpler-env/SimplerEnv}} presents the simulated
evaluations as ``valuable quantitative metrics for checkpoint selection'' \citep{simpler}. Later work selects checkpoints by simulated success on the strength of an
observed sim-to-real correlation \citep{shi2026beyond}, and \citet{wang2026recipe} write that a lower
MMRV makes a simulator less likely to mislead model selection. None of these sources states how much
a choice can lose. This paper does. For routing, the loss that the MMRV allows falls linearly with its
value. For deployment, even an MMRV of zero leaves a loss that only the outcome table can bound.

The same reading runs through other fields. BLEU was introduced as an understudy ``which substitutes
for'' human judges, on the strength of a correlation of $.99$ over five systems
\citep{papineni2002bleu}. That value still allows a loss of about a fifth of the human scale. AlpacaEval
reports a Spearman correlation of $.98$ with Chatbot Arena over $38$ models \citep{dubois2024lc}, and
at that size a rank correlation guarantees nothing. RewardBench~2 offers itself as a guide for picking a
reward model, with a correlation of $.87$ over $113$ of them \citep{malik2026rewardbench2}. With $113$
candidates, any correlation below $.987$ is vacuous. PPE expects improvements on its benchmark to
carry over to downstream gains on the strength of a correlation of $.80$ over nine reward models
\citep{frick2025ppe}, below the value at which a correlation over nine stops being vacuous. The authors
often know the limits. RewardBench~2 itself warns that the top-scoring reward model does not ensure a
good model after RLHF. What has been missing is the size of the gap.

We turn a published agreement statistic into a \emph{guarantee}: the largest regret, the real outcome
lost against the best available choice, that any evaluator with that value could cause. It is a worst
case over evaluators, not a probability. We compute it in two ways. The \emph{value-only bound} uses
only the published value and $K$, and so covers every possible outcome table. The \emph{instance bound} uses the published outcome table and is exact for it. We apply both to published
validations from robotics, machine translation, language-model pretraining, reward modelling and
architecture search. We find three things that reporting a correlation hides.
\begin{itemize}[leftmargin=*,topsep=2pt,itemsep=2pt,parsep=0pt]
\item \textbf{With many policies, a correlation is vacuous.} Unless $1-r^2<2.92/K$, its
value-only bound is the largest possible regret (Theorem~\ref{thm:everyboard}). At $r=.9$, it is vacuous from exactly fifteen policies on. In eleven robot-evaluation papers, none of the $51$ published
correlations guarantees routing within five points. Even the highest, $.995$, cannot rule
out a loss of $14.4$ points (Sec.~\ref{sec:exchange}).
\item \textbf{The better the evaluator, the more a correlation overstates its regret.} For Gaussian errors of size $\sigma$, the correlation's value-only bound falls like $\sigma$, but the expected regret and the MMRV's expected guarantee are at most of order $\sigma^2$ (Theorem~\ref{thm:noise}), since that guarantee is linear in the MMRV. On every published table we examine, the MMRV's instance bound for routing is never looser than the correlation's.
\item \textbf{From its value alone, no statistic that only compares policies within tasks can certify which policy to deploy}, even at its
perfect value: on each outcome table its guarantee has a floor set by that table. This holds for the MMRV, for pairwise accuracy and for averages of within-task
correlations (Theorem~\ref{thm:levels}). On WorldGym, a perfect MMRV still leaves $52$ points of
deployment regret open, while the pooled correlation, $.785$ (printed as $.78$), rules out any.
\end{itemize}
Unlike earlier regret bounds (Sec.~\ref{sec:related}), ours need no model of the errors
and hold for finite tables on any bounded scale. Their constants are explicit: exact for the MMRV and
tight to first order for the correlation. We therefore recommend publishing both tables, or else the MMRV with each policy's real and evaluator mean (Sec.~\ref{sec:q3}). We release a library (\texttt{evalceil};
Sec.~\ref{sec:related}, Reproducibility) that turns a reported statistic, or a pair of published tables, into its
guarantees.

%% file: setting.tex
\section{Setting}\label{sec:setting}

\paragraph{Tables and decisions.} The \emph{outcome table} $p\in[0,1]^{K\times T}$ holds the
published real outcomes: $p_{it}$ is the outcome of policy $i$ on task $t$. Outcomes are success
rates, or human quality scores and accuracies mapped to $[0,1]$; every guarantee below transfers to
any bounded scale, multiplied by its width. The \emph{score table} $y\in\mathbb R^{K\times T}$ holds
the evaluator's scores. On task $t$, the sorted outcomes are $p_{(1)t}\ge\cdots\ge p_{(K)t}$, and the
evaluator \emph{picks} the policy it scores highest, $a(t)$. Ties are broken against the evaluator, toward the worse policy. We study three decisions. \emph{Routing}
(per-task selection) uses the pick on every task, as systems that route tasks to policies do
\citep{ong2025routellm}. Its regret is
\begin{equation}\label{eq:regret}
R(a)\;=\;\frac1T\sum_{t=1}^{T}\bigl(p_{(1)t}-p_{a(t)t}\bigr).
\end{equation}
This is per-instance algorithm selection \citep{rice1976algorithm,kerschke2019survey}, where choosing the per-task optimum is the \emph{virtual best solver} and using the best-mean policy everywhere the \emph{single best solver}; a selector is scored there by how much of the gap between the two it closes. \emph{Deployment} (single-policy selection) puts into use the policy with the highest mean score. Its regret is the best mean
outcome minus the mean outcome of the deployed policy. \emph{Top-$k$ screening} tests the $k<K$
policies with the highest mean scores for real and deploys the best of them. Regret is a fraction of the
outcome's range, so \emph{five points} is $\varepsilon=.05$.

\paragraph{Statistics.} The \emph{correlation} $r$ is the Pearson correlation between $y$ and $p$
over all $K\times T$ entries, also called the \emph{pooled} correlation. We write $\rho=1-r^2$, the
share of variance it leaves unexplained. The \emph{mean maximum rank violation} (MMRV), introduced by
SIMPLER \citep{simpler}, looks at each policy $i$ on each task $t$. Among the policies $j$ whose order
relative to $i$ the evaluator gets wrong, it takes the largest real gap $\lvert p_{it}-p_{jt}\rvert$.
It averages these maxima:
\begin{equation}\label{eq:mmrv}
\mathrm{MMRV}(y,p)=\frac{1}{KT}\sum_{t=1}^{T}\sum_{i=1}^{K}\ \max_{j\,:\,[y_{it}>y_{jt}]\neq[p_{it}>p_{jt}]}\lvert p_{it}-p_{jt}\rvert .
\end{equation}
An empty maximum is zero, and a tie in scores counts as a violation for the policy with the higher real outcome only, the mirror of SIMPLER's rule (App.~\ref{app:summaries}).
In Figure~\ref{fig:example}, Evaluator~1 swaps one pair, B and C on the second task, whose real gap is
$.2$. Both policies then have $.2$ as their maximum in~\eqref{eq:mmrv}, so the MMRV is $(.2+.2)/9=.044$. Evaluator~2 makes one such swap
on every task, so its MMRV is $6\times.2/9=.133$.

\paragraph{Guarantees.} The \emph{guarantee} of a published value is the largest regret of any score
table whose statistic is at least as good as that value. The \emph{instance bound} takes the
published outcome table as given. The \emph{value-only bound} also maximises over all outcome tables
with $K$ policies, so it needs only the value and $K$; it is never smaller than the instance bound. A
statistic \emph{certifies} a decision when its guarantee is at most a tolerance $\varepsilon$, five
points unless stated. It is \emph{vacuous} when its guarantee is the largest regret possible:
$1$ over all tables, and the regret of the worst possible choice on a given one. We write
$\varepsilon^\star(r)$ for the instance bound of a correlation $r$ for routing. In
Figure~\ref{fig:example}, the correlation of $.954$ has a value-only bound of $.31$ and an instance bound of $.200$, which Evaluator~2 attains.

\paragraph{From a correlation to a guarantee.} Let $S=\sum_{i,t}(p_{it}-\bar p)^2$ be the total sum of squares of $p$ about its grand mean $\bar p$. A correlation is unchanged by a positive affine rescaling of the scores. For $r>0$ the best such rescaling is the least-squares fit of $p$ on $y$, whose residual sum of squares is $S(1-r^2)$. Every wrong pick uses up some of this budget of squared error. The worst evaluator spends it on the wrong picks that buy the most regret per unit of squared
error. For a choice $a$ (one policy per task), let $\mathcal C_a=\{y: y_{a(t)t}\ge y_{it}\ \forall
i,t\}$ be the score tables under which the evaluator makes that choice. Let $d(a)=\min_{y\in\mathcal
C_a}\lVert y-p\rVert_F^2$ be the squared distance from $p$ to that set. This distance is a sum over tasks, and each term is an isotonic regression under a simple tree order rooted at the pick \citep{robertson1988}.

\begin{proposition}\label{prop:cone}
For $K\ge2$, $S>0$ and any choice $a$, the largest correlation that a nonconstant score table in $\mathcal C_a$ can have with $p$ is at most $\sqrt{1-d(a)/S}$, with equality whenever $d(a)<S$.
\end{proposition}

An evaluator with correlation at least $r$ can therefore make exactly the choices with
$d(a)\le S(1-r^2)$ (App.~\ref{app:body} treats a boundary case), and
\begin{equation}\label{eq:exchange}
\varepsilon^\star(r)\;=\;\max\bigl\{R(a)\ :\ d(a)\le S(1-r^2)\bigr\},\qquad r\in[0,1].
\end{equation}
Problem~\eqref{eq:exchange} is a multiple-choice knapsack \citep{kellerer2004knapsack}: each task is a class, each policy an item, with its term of $d(a)$ as weight and its regret on the task as profit. We solve it exactly by dynamic programming in rational arithmetic. The worst
evaluator ties each wrong pick with the best policy, but ties are not essential: for all but finitely
many $r$, an evaluator without ties attains $\varepsilon^\star(r)$. Proposition~\ref{prop:cone} rests on Moreau's decomposition into a closed convex cone and its polar \citep{moreau1962decomposition}. The same construction covers
deployment and top-$k$ screening (App.~\ref{app:coarse}). All proofs are in the appendix.

%% file: report.tex
\section{What a reported number guarantees}\label{sec:report}

A correlation of $.9$ guarantees nothing once fifteen policies are compared. Most validation papers publish one number, a correlation or an MMRV, together with $K$, and leave out
the tables. This section gives the value-only bound of each statistic: what a reader can conclude from
that number, whatever tables lie behind it.

\begin{proposition}\label{prop:rank}
Let the score table $y$ have $\mathrm{MMRV}(y,p)\le v$, and let the choice $a$ take a highest-scoring
policy on each task. Then \emph{(i)} $R(a)\le Kv$; and \emph{(ii)} if every task has a unique top score, $R(a)\le\tfrac{K}{2}v$.
\end{proposition}

When the pick on a task is wrong, the best policy's term in
\eqref{eq:mmrv} is at least the regret on that task. With a unique top score, so is the pick's term.
Since \eqref{eq:mmrv} averages $K$ terms per task, $R(a)\le Kv$, or $\tfrac K2v$. Both bounds are
attained (Theorem~\ref{thm:everyboard}(i)). They also hold under SIMPLER's tie rule. In Figure~\ref{fig:example}, every top score is unique,
so the value-only bounds of the two MMRVs are $\tfrac32\times.044=.067$ and
$\tfrac32\times.133=.200$: exactly the two evaluators' losses. Each term of \eqref{eq:mmrv} is the real gap of a
misranking, the unit in which a decision loses, so the MMRV's guarantee is linear in its value.

\paragraph{A correlation leaves room for cheap mistakes.} A wrong pick behind a gap $g$ costs
squared error of order $g^2$ but loses $g$. The budget $S(1-r^2)$ of Prop.~\ref{prop:cone} therefore
buys the most regret when it is spread over all $T$ tasks. A gap $g$ on every task costs at least $Tg^2/2$, which cannot exceed $S(1-r^2)\le KT\rho/4$, so $g\le\sqrt{K\rho/2}$, the rate of Theorem~\ref{thm:everyboard}(ii).

\begin{theorem}[worst-case per-task regret given a published statistic]\label{thm:everyboard}
Fix $K\ge2$. Over all outcome tables with $K$ policies and all score tables:
\begin{enumerate}[label=(\roman*),leftmargin=2em,topsep=1pt,itemsep=1pt,parsep=0pt]
\item the worst regret with an MMRV of at most $v$ is $\min\{1,Kv\}$, and $\min\{1,Kv/2\}$ when the evaluator's top policy is unique on every task, so the bounds of Prop.~\ref{prop:rank} are tight;
\item the worst regret with a correlation of at least $r\in[0,1)$ is at most $B_K(\rho)\le\sqrt{K\rho/2}$, with $B_K$ explicit (App.~\ref{app:universal}). The bound is tight for $K=2$, and for $K=3$ when $\rho\le\frac14$.
For $K\ge4$ the worst regret is at least $\min\{1,\underline B_K(\rho)\}$, with $\underline B_K$ equal to $B_K$ to
first order as $r\to1$;
\item for $K\ge3$ there is a vacuity threshold $r_{\mathrm{vac}}(K)$: the worst regret is $1$, so the correlation is vacuous, exactly when $r\le r_{\mathrm{vac}}(K)$; and $1-r_{\mathrm{vac}}(K)^2<(3+2\sqrt2)/(2K)<2.92/K$;
\item with a Kendall coefficient of at least $\tau$, or a Spearman coefficient of at least $r_{\mathrm S}$, pooled over all $KT$ entries, with no ties, the worst regret is $\min\{1,K\sqrt{(1-\tau)/2}\}$, or $\min\{1,K((1-r_{\mathrm S})/4)^{1/3}\}$.
\end{enumerate}
\end{theorem}

\begin{table}[t]
\centering\small\setlength{\belowcaptionskip}{3pt}
\caption{Value-only bounds: the worst per-task regret allowed by each published statistic over all outcome tables with $K$ policies (Theorem~\ref{thm:everyboard} and App.~\ref{app:separable}; every bound is capped at $1$), and the value each must reach to guarantee five points with five policies.}
\label{tab:rates}
\begin{tabular}{@{}lll@{}}
\toprule
published statistic & worst per-task regret & five points, $K=5$\\
\midrule
MMRV $v$ & $Kv$, or $Kv/2$ with a unique top-scored policy (tight) & $v\le.02$ (unique top-scored)\\
Pearson $r$ & at most $B_K(\rho)\le\sqrt{K\rho/2}$, $\rho=1-r^2$ & $r\ge.99949$\\
Kendall $\tau$ & $K\sqrt{(1-\tau)/2}$ & $\tau\ge.9998$\\
Spearman $r_{\mathrm S}$ & $K\bigl((1-r_{\mathrm S})/4\bigr)^{1/3}$ & $r_{\mathrm S}\ge.999996$\\
pairwise accuracy $A$ & $(1-A)K(K-1)/2$ (supremum) & $A\ge.995$\\
\bottomrule
\end{tabular}
\end{table}

Rank coefficients do no better than Pearson's: their guarantees grow linearly in $K$, not as $\sqrt K$, and scale as the square root (Kendall)
and the cube root (Spearman) of the distance from a perfect value. Two consequences matter in
practice.

\paragraph{More policies need a higher correlation.} At or below $r_{\mathrm{vac}}(K)$, a correlation is
vacuous, and this threshold approaches one as $K$ grows. Rounded down, $r_{\mathrm{vac}}$ is $.633$ for
four policies, $.707$ for five and $.853$ for ten. Put the other way, a correlation $r$ is non-vacuous only for fewer than $2.92/(1-r^2)$ policies; at $r=.9$, for at most fourteen
(Theorem~\ref{thm:universal} in App.~\ref{app:universal}).  With five policies, a correlation guarantees five points on every
outcome table only above a threshold between $.99947$ and $.99949$ (Table~\ref{tab:needed} gives other $K$ and tolerances). The MMRV needs only $v\le.02$ when top scores are unique.

\paragraph{Statistics computed within tasks.} Several papers average a correlation $c_t$ computed
within each task instead of pooling it. For routing, this average obeys the same
square-root law. With an average of at least $\bar c$, the worst regret is at most
$\sqrt{2\Sigma_K(1-\bar c_+^2)}$, where $\Sigma_K=\lfloor K/2\rfloor\lceil K/2\rceil/K$ and
$\bar c_+=\max\{\bar c,0\}$; for $K\ge3$, this is tight to first order as $\bar c\to1$
(Prop.~\ref{prop:within}). Pairwise accuracy $A$, the share of policy pairs ordered correctly within tasks, is used to rank machine-translation metrics and language-model judges. Without ties, $2A-1$ is the task average of Kendall's $\tau$. Its guarantee is linear in $1-A$, like the MMRV's, but it counts misordered pairs whatever their gaps, so its constant is $K(K-1)/2$ rather than $K$ (Table~\ref{tab:rates}). Every statistic that sums per-task terms has an exact instance bound, computed by one knapsack over tasks (App.~\ref{app:separable}).

Deployment is another matter. Rescaling one task's scores by a positive factor leaves every $c_t$, the MMRV and $A$ unchanged, but it can change which policy has the highest mean. So no within-task statistic, from its value alone, bounds the regret of deployment or of top-$k$ screening (Sec.~\ref{sec:exchange}).

\begin{figure}[t]
\centering\includegraphics[width=0.55\textwidth]{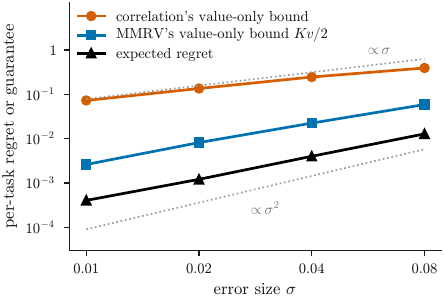}
\caption{\textbf{Random errors of size $\sigma$} on the nine corpus outcome tables ($4{,}000$ draws each, means). Each halving of $\sigma$ divides the correlation's value-only bound (the lower end of Theorem~\ref{thm:everyboard}(ii)) by $1.75$ on average, but the MMRV's guarantee $Kv/2$ by $2.8$ and the regret by $3.2$ (Theorem~\ref{thm:noise} bounds both by order $\sigma^2$).}
\label{fig:noise}
\end{figure}
\paragraph{On the scale of the evaluator's errors.} The $\sqrt{1-r^2}$ rate is not an artefact of squaring $r$. Measured against the size $\sigma$ of the evaluator's errors, the correlation still guarantees only order $\sigma$, while the regret is of order $\sigma^2$. In this model $r^2$ is close to $S/(S+KT\sigma^2)$, the reliability ratio of classical measurement error.

\begin{theorem}[guarantees on the scale of the errors]\label{thm:noise}
Let $S>0$ and $KT\ge3$, and let $y=p+\sigma Z$, where $Z$ has independent standard normal entries.
\emph{(i)} As $\sigma\to0$, the correlation's value-only bound is $\sigma\sqrt{KW/2}\,(1+o(1))$
almost surely, where $SW$ has a chi-squared distribution with $KT-2$ degrees of freedom (App.~\ref{app:noise}), so $\E W=(KT-2)/S$.
\emph{(ii)} With $\gamma$ the smallest positive gap between two policies on a task and $N\ge1$ the number
of (task, pair of policies) combinations with a positive gap, $\Pr[\mathrm{MMRV}>0]\le\frac
N2e^{-\gamma^2/(4\sigma^2)}$. Where the MMRV is $0$, the regret is $0$. So the regret and the MMRV's
guarantee $\frac K2\mathrm{MMRV}$ have expectation at most $2N\sigma^2/(e\gamma^2)$ and $KN\sigma^2/(e\gamma^2)$.
\emph{(iii)} If, in addition, the outcomes on each task are drawn independently, and independently of
$Z$, from densities bounded by $f$, then the MMRV's guarantee and the regret
have expectation at most $\frac{K(K-1)}2f\sigma^2$.
\end{theorem}

The regret and the MMRV's guarantee are exponentially small once $\sigma$ is well below the smallest gap. The contrast concerns what the value alone can say: with the table, the correlation's instance bound also
vanishes once $\sigma$ is small enough. On the nine corpus outcome tables (Sec.~\ref{sec:exchange}), as $\sigma$ falls from $.08$ to $.01$, the correlation's value-only bound rises from $31$ to $180$ times the regret, and
the MMRV's from $4.6$ to $6.5$ times (Figure~\ref{fig:noise}).

%% file: exchange.tex
\section{What published validations guarantee}\label{sec:exchange}\suppressfloats[t]

We apply the guarantees to three bodies of evidence, summarised in Table~\ref{tab:sources} and
described in Apps.~\ref{app:corpus} and~\ref{app:audits}. Throughout, a \emph{validation} is one
evaluator scored against one outcome table. The \emph{corpus} holds the eleven validations that
publish both tables, on nine outcome tables from six papers: eight robot evaluators
(WorldGym \citep{worldgym}, Ctrl-World \citep{ctrlworld}, SIMPLER \citep{simpler}, AutoEval
\citep{autoeval}), two web-agent judges from AgentRewardBench \citep{agentrewardbench} and one
translation metric \citep{kocmi-etal-2023-findings,popovic2015chrf}. Only WorldGym publishes the pooled
correlation, so we recompute it for all eleven. The \emph{reported statistics} are the agreement statistics
printed for $52$ validations in eleven robot-evaluation papers, mostly without the tables behind them. The
four \emph{audits} score many evaluators on each outcome table; in PPE, a
policy is a sample index among $32$ answers to an item. For each audit, a criteria file fixed the table construction and the hypotheses before any evaluator score was read. The files are not publicly time-stamped (App.~\ref{app:audits}). Of the $25$ registered hypothesis parts,
$5$ follow closely from Theorem~\ref{thm:everyboard} at the audit's $K$ (a grouping made after the audits), and all held. Of the $20$ others, $14$ held and $6$ failed.

\begin{table}[h]
\centering\footnotesize\setlength{\belowcaptionskip}{3pt}
\caption{The three bodies of evidence (Apps.~\ref{app:corpus},~\ref{app:audits}). \emph{Published}: what the source prints.}
\label{tab:sources}
\setlength{\tabcolsep}{3.5pt}\begin{tabular}{@{}llllrrr@{}}
\toprule
& source & evaluators & published & validations & $K$ & $T$\\
\midrule
corpus & six papers & robot evaluators, judges, a metric & both tables & $11$ & $3$--$9$ & $4$--$17$\\
reported statistics & eleven robot papers & robot evaluators & $r$ and/or MMRV & $52$ & $2$--$8$ & mostly $1$\\
audits & WMT22--25 & translation metrics & both tables & $702$ & $10$--$20$ & $2$--$5$\\
 & DataDecide & small-model proxies & both tables & $13$ & $25$ & $10$\\
 & PPE & reward models & both tables & $70$ & $32$ & $489$--$510$\\
 & NAS-Bench-201 & zero-cost proxies & both tables & $13$ & $15{,}625$ & $3$\\
\bottomrule
\end{tabular}
\end{table}

\begin{figure}[b]
\centering\includegraphics[width=\linewidth]{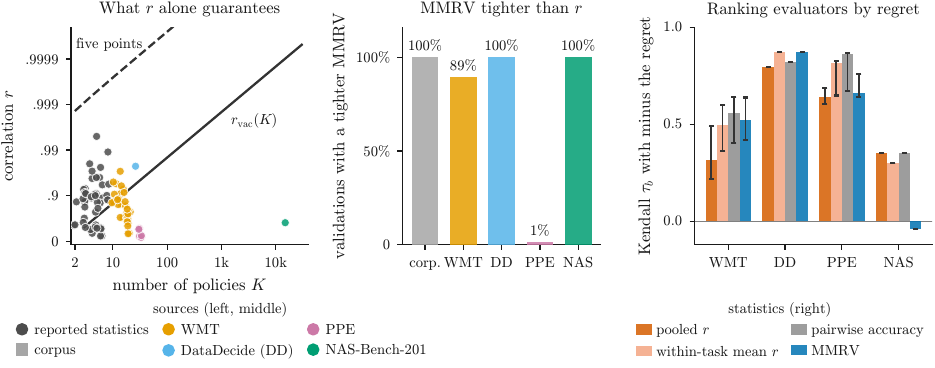}
\caption{\textbf{Correlations guarantee little; the MMRV mostly tracks the regret.} Left: for each outcome table, its highest correlation $r$ (as printed in the robot papers, pooled or on a single task; pooled and recomputed over each audit table's evaluators) against its number of policies $K$, jittered horizontally; correlations at or below zero are not drawn. At or below the solid curve $r_{\mathrm{vac}}(K)$, $r$ alone is vacuous; above the dashed curve it guarantees routing within five points (Theorem~\ref{thm:everyboard}). Middle: the share of validations on which the MMRV's instance bound for routing is strictly tighter than the correlation's; on the others the two are equal. Right: Kendall's $\tau_b$ between each statistic (signed so that higher is better; the within-task mean correlation is exploratory) and minus the evaluators' realised regret, median over each source's tables, with the interquartile range over WMT's $32$ tables and PPE's $5$.}
\label{fig:audits}
\end{figure}

\paragraph{From the value alone, correlations guarantee little.} On the audits, $766$ of the $798$ correlations are at or below $r_{\mathrm{vac}}(K)$ and hence vacuous (Sec.~\ref{sec:report}; Figure~\ref{fig:audits}, left): $673$ of $702$ on WMT, $10$ of $13$ on DataDecide, and all on PPE and NAS. The robot papers, with at most eight policies, are less exposed. Yet none of
their $51$ reported correlations guarantees five points. The highest, $.995$, is printed for a single task and
still leaves open a regret of $.144$ ($14.4$ points). The MMRV guarantees five points
on $7$ of the $48$ robot validations that report it. Where a validation reports both statistics, the
MMRV's value-only bound is strictly tighter on $32$ of $47$, and the correlation's on $4$. The robot papers'
averages of within-task correlations also fall short. Each leaves open a per-task regret of at least
$.40$, and three are vacuous (App.~\ref{app:census}).

\paragraph{With the tables, coarse decisions are sometimes certified, fine ones are not.} Deployment
and top-$k$ screening are coarse decisions: they use only the order of the evaluator's policy means. Routing uses its order within every task. On the corpus, the correlation's instance bound certifies deployment within five
points on two validations, and a top-2 screen on six (Table~\ref{tab:decisions}). It certifies
routing on none. On ten validations it cannot even rule out doing worse than the \emph{single best policy}, the one with the best mean real outcome, used on every task.  This is not only a worst case.
On four validations, the published evaluator itself did worse per task than the single best policy:
SIMPLER with scores averaged over visual variants loses $19.7$ points per task, against $9.7$, and
SIMPLER scored against AutoEval's real trials $41.5$ against $25.5$
(Table~\ref{tab:eleven}). A correlation can even be earned by task difficulty alone. For
AgentRewardBench's rule-based checker, scores that only knew each task's average success rate would reach at least the same $r$ (their correlation is the correlation ratio of $p$ on the tasks), and the correlation's instance bound is the largest regret possible
(Cor.~\ref{cor:vacuous}). The audits repeat the pattern at scale. DataDecide's 530M models correlate
with the 1B table at $.977$, above $r_{\mathrm{vac}}(25)$, yet the correlation's instance bound is
the largest regret possible (App.~\ref{app:datadecide}). On PPE, every reward model's statistics
guarantee only the largest possible regret, except one MMRV (App.~\ref{app:ppe}).

\input{table_eleven}

\paragraph{With the tables, the MMRV is never looser.} For routing, the MMRV's instance bound is tighter than the correlation's on all eleven corpus validations. On the ten where it is
positive, it is tighter by a factor of $1.15$ to $6.33$, with median $2.9$ (Table~\ref{tab:eleven}). This is a regularity of the published tables, not a theorem: App.~\ref{app:separable} gives a table on which the order reverses. On the audits, the
MMRV's instance bound is below the correlation's on $628$ WMT validations, all $13$ DataDecide sizes,
all $13$ NAS proxies and one PPE validation, and equal elsewhere (Figure~\ref{fig:audits}, middle).
Pairwise accuracy, which WMT and PPE report, counts misordered pairs rather than their gaps. Its instance bound is below the correlation's on only $59$ of the $702$ WMT validations.

\input{table_decisions}

\paragraph{Deployment needs the levels across tasks.} An evaluator can order every policy correctly on every task and still deploy the worst one. Call a policy \emph{undominated} (Pareto-optimal across tasks) when no other policy is at least as good on every task and better on one. Let $\Phi(p)$ be the best policy mean
minus the lowest mean of an undominated policy.

\begin{theorem}[deployment: levels against orders]\label{thm:levels}
Let a statistic depend on the score table only through its weak order within each task. At every published value that the perfect scores $y=p$ also attain, its deployment guarantee is at least
$\Phi(p)$ (proof in App.~\ref{app:levels}).
\end{theorem}

The MMRV and pairwise accuracy are such statistics. They do not penalise an evaluator for reordering
policies that the outcomes tie, so at a perfect value their guarantee is exactly a floor at least as large, $\Phi_0(p)$, taken over the weakly Pareto-optimal policies, those that no other beats on every task (App.~\ref{app:levels}). In Figure~\ref{fig:example},
$\Phi_0=.200$, and Evaluator~3 attains it. On WMT23, $\Phi_0=.045$, so a perfect MMRV certifies
deployment. On WorldGym, $\Phi_0=.521$, whereas the pooled correlation, $.785$, constrains the
levels across tasks and guarantees zero deployment regret. On Ctrl-World's instruction table, $\Phi=0$,
yet an evaluator with an MMRV of $0$ can deploy a policy that loses $\Phi_0=.30$. Counted by outcome table, $\Phi_0$ exceeds five points on $8$ of the $9$ in
the corpus (Table~\ref{tab:sepship}). On the $33$ WMT tables (the $32$ audited and one that no
metric scores in full), even the smaller floor $\Phi(p)$ exceeds five points on $10$. A validation must therefore also
report each policy's mean (Sec.~\ref{sec:q3}).

\paragraph{Which statistic tracks the regret.} A single validation has one evaluator and so one
realised regret, so it cannot show whether a statistic ranks evaluators by their regret. The audits
can. Over the WMT tables, the median Kendall $\tau_b$ between a statistic and minus the realised
regret is $.32$ for the correlation and $.52$ for the MMRV; the MMRV's is higher on $27$ of the $32$
tables, and a bootstrap over tables ($10{,}000$ resamples) puts the difference in $[.08,.30]$ (95\% percentile interval). The values are $.79$ against $.87$ on DataDecide, and $.64$ against $.66$
on PPE (Figure~\ref{fig:audits}, right). Pairwise accuracy does as well or better on WMT and PPE
($.56$ and $.86$), but its worst-case guarantee is looser. The average of within-task correlations,
which several robot papers report, also beats the pooled one on three sources ($.49$ on WMT, $.87$ on DataDecide, $.82$ on PPE) but not on NAS ($.30$ against $.35$; exploratory, not registered; App.~\ref{app:wmtaudit}).
NAS is the exception overall: the MMRV orders its $13$ proxies' regret worse than the correlation
($-.04$ against $.35$), while Precision@100, which scores top-$k$ screening directly, orders them by screening regret with $\tau_b=.79$ (App.~\ref{app:nas}). One registered hypothesis cuts against the MMRV. On WMT, a metric's MMRV guarantee from one
year bounds its regret on the next year's table in only $39$ of $55$ cases, against $53$ for the
looser correlation's (App.~\ref{app:wmtaudit}).

\paragraph{The published trial counts limit every robot guarantee.} Four real experiments report how
many trials each cell averages: WorldGym, SIMPLER's Google Robot and WidowX, and AutoEval. Score $200$
real tables, resampled at those counts, with an evaluator whose scores are the published rates. Its
correlation certifies five points on $0$, $0$, $30$ and $31$ of them, its MMRV on $198$, $190$, $191$
and $199$ (App.~\ref{app:evidence}). If each true rate may lie anywhere in its $95\%$ binomial interval,
no choice is guaranteed within five points on any of the four tables, even by a perfect evaluator; only
the four-task subset of AutoEval's SIMPLER comparison escapes, at $.048$ (App.~\ref{app:hinge}). These
validations are limited by their trials, whatever the evaluator.

%% file: table_eleven.tex
\begin{table}[t]
\centering\small\setlength{\tabcolsep}{4pt}\setlength{\belowcaptionskip}{3pt}
\caption{The eleven validations that publish both tables. Guarantees are instance bounds for routing: $\varepsilon^\star(r)$ for the pooled correlation $r$, and the MMRV's instance bound (App.~\ref{app:separable}). A tie in the scores counts against the policy with the higher real rate, Eq.~\eqref{eq:mmrv}. The MMRV's worst case is taken over evaluators with a unique top-scored policy on every task when the published one has it, and over any pattern of ties otherwise. Rows: SIMPLER on a matched scene or averaged over visual variants; Ctrl-World's two outcome tables; AutoEval's autonomous scorer against supervised outcomes, SIMPLER against AutoEval's real trials. \emph{Realised}: the published evaluator's regret, ties broken against it; \emph{single best}: the per-task regret of using the best-mean policy on every task.}
\label{tab:eleven}
\begin{tabular}{@{}lrrcccccc@{}}
\toprule
 & & & & & \multicolumn{2}{c}{guarantee of} & \multicolumn{2}{c}{regret}\\
\cmidrule(lr){6-7}\cmidrule(l){8-9}
evaluator and outcome table & $K$ & $T$ & $r$ & MMRV & $r$ & MMRV & realised & single best\\
\midrule
WorldGym & $3$ & $17$ & $.785$ & $.0480$ & $.447$ & $.071$ & $.012$ & $.024$ \\
SIMPLER Google, visual match & $6$ & $4$ & $.941$ & $.0494$ & $.335$ & $.128$ & $.017$ & $.097$ \\
SIMPLER Google, variant agg. & $6$ & $4$ & $.834$ & $.1661$ & $.498$ & $.310$ & $.197$ & $.097$ \\
SIMPLER WidowX & $3$ & $4$ & $.841$ & $.0278$ & $.165$ & $.042$ & $.042$ & $.042$ \\
Ctrl-World, instruction & $3$ & $7$ & $.968$ & $.0000$ & $.114$ & $.000$ & $.000$ & $.000$ \\
Ctrl-World, success & $3$ & $7$ & $.832$ & $.0190$ & $.293$ & $.057$ & $.007$ & $.007$ \\
AutoEval, autonomous & $6$ & $5$ & $.992$ & $.0147$ & $.096$ & $.044$ & $.000$ & $.236$ \\
AutoEval, SIMPLER & $6$ & $4$ & $.450$ & $.2067$ & $.870$ & $.525$ & $.415$ & $.255$ \\
AgentRewardBench, judge & $4$ & $4$ & $.965$ & $.0286$ & $.146$ & $.044$ & $.000$ & $.009$ \\
AgentRewardBench, rule & $4$ & $4$ & $.931$ & $.0751$ & $.189$ & $.164$ & $.071$ & $.009$ \\
WMT23 chrF & $9$ & $8$ & $.781$ & $.0196$ & $.132$ & $.042$ & $.011$ & $.006$ \\
\bottomrule
\end{tabular}
\end{table}

%% file: table_decisions.tex
\begin{table}[t]
\centering\footnotesize\setlength{\belowcaptionskip}{3pt}\setlength{\tabcolsep}{4pt}
\caption{Deployment and top-$k$ screening on the eleven validations. \emph{Guarantee}: the largest regret any evaluator with at least the validation's pooled correlation can cause (App.~\ref{app:coarse}); \emph{realised}: the published evaluator's regret. $\Phi(p)$, $\Phi_0(p)$: the floors of Theorem~\ref{thm:levels} and App.~\ref{app:levels}. $k^\star$: the smallest $k$ whose top-$k$ screen the correlation certifies within five points. \emph{Saved}: the share of hardware trials the screen saves when the deployed policy must be within five points of the best with probability $.95$ (App.~\ref{app:practice}).}
\label{tab:decisions}\label{tab:sepship}
\begin{tabular}{@{}lccccccccc@{}}
\toprule
 & & \multicolumn{4}{c}{deployment} & \multicolumn{2}{c}{top-2 screen} & & \\
\cmidrule(lr){3-6}\cmidrule(lr){7-8}
evaluator and outcome table & $r$ & guarantee & realised & $\Phi(p)$ & $\Phi_0(p)$ & guarantee & realised & $k^\star$ & saved\\
\midrule
WorldGym & $.785$ & $.000$ & $.000$ & $.521$ & $.521$ & $.000$ & $.000$ & $1$ & $100\%$ \\
SIMPLER Google, visual match & $.941$ & $.087$ & $.000$ & $.087$ & $.087$ & $.074$ & $.000$ & $3$ & $41\%$ \\
SIMPLER Google, variant agg. & $.834$ & $.391$ & $.087$ & $.087$ & $.087$ & $.087$ & $.000$ & $4$ & $21\%$ \\
SIMPLER WidowX & $.841$ & $.061$ & $.000$ & $.061$ & $.061$ & $.000$ & $.000$ & $2$ & $46\%$ \\
Ctrl-World, instruction & $.968$ & $.000$ & $.000$ & $.000$ & $.300$ & $.000$ & $.000$ & $1$ & $100\%$ \\
Ctrl-World, success & $.832$ & $.279$ & $.000$ & $.279$ & $.279$ & $.000$ & $.000$ & $2$ & $46\%$ \\
AutoEval, autonomous & $.992$ & $.112$ & $.004$ & $.112$ & $.408$ & $.004$ & $.000$ & $2$ & $74\%$ \\
AutoEval, SIMPLER & $.450$ & $.615$ & $.000$ & $.180$ & $.180$ & $.615$ & $.000$ & $5$ & $0\%$ \\
\addlinespace[1.5pt]
AgentRewardBench, judge & $.965$ & $.155$ & $.000$ & $.070$ & $.070$ & $.070$ & $.000$ & $3$ & $32\%$ \\
AgentRewardBench, rule & $.931$ & $.155$ & $.125$ & $.070$ & $.070$ & $.125$ & $.000$ & $4$ & $0\%$ \\
\addlinespace[1.5pt]
WMT23 chrF & $.781$ & $.119$ & $.007$ & $.045$ & $.045$ & $.049$ & $.007$ & $2$ & $4\%$ \\
\bottomrule
\end{tabular}
\end{table}

%% file: card.tex
\section{What to publish, and how to use an evaluator}\label{sec:q3}

\paragraph{The validation card.} No single number can tell a reader both which policy to use on each task and which policy to deploy (Prop.~\ref{prop:noscalar}). A benchmark can let every reader compute the guarantee for their own
decision without rerunning anything. The card (Figure~\ref{fig:card}) lists what that takes. The note
after each item says how many of our sources already report it. Item~2 asks for two kinds of number because one number cannot track both decisions.

\begin{figure}[t]
\noindent\fbox{\begin{minipage}{\dimexpr\linewidth-2\fboxsep-2\fboxrule\relax}\small
\textbf{Evaluator validation card}, for an outcome table of $K$ policies and $T$ tasks.
\begin{enumerate}[leftmargin=*,topsep=1pt,itemsep=1pt,parsep=0pt]
\item Both tables, with the number of trials or judgements behind each outcome, and the realised regret of each decision. They give every instance bound: the correlation's, Eq.~\eqref{eq:exchange}, and that of every statistic that sums per-task terms (App.~\ref{app:separable}). \emph{(Already reported: both tables by the six corpus papers and the WMT metrics tasks; trial counts by four real experiments.)}
\item Failing the tables: the MMRV with $K$, and whether the evaluator's top policy is unique on every task (Prop.~\ref{prop:rank}); each policy's real and evaluator mean, which deployment and top-$k$ screening need and the MMRV does not give; and the WRV and $\mathrm{WRV}_{\mathrm{dep}}$ (defined below), the least bounds valid for every reader's candidate subset (Theorem~\ref{thm:menus}). A correlation, if reported, should say whether it is pooled or within each task, and never stand alone. \emph{(MMRV: $48$ of the $52$ robot validations we collected.)}
\item The decision the evaluator is validated for, and its scope: the tasks and policies the guarantee covers. \emph{(No source states the decision in full.)}
\end{enumerate}
\end{minipage}}
\caption{\textbf{Evaluator validation card.} What a validation should publish.}
\label{fig:card}
\end{figure}

\begin{proposition}[one number cannot track both decisions]\label{prop:noscalar}
Fix $K\ge2$. No statistic $s$ of a pair of tables has monotone functions $G_1$ and $G_2$, each
nondecreasing or nonincreasing, such that $G_1(s)$ is the realised per-task regret and $G_2(s)$
the realised deployment regret on every pair of tables.
\end{proposition}

Routing needs the order within each task, and deployment needs the levels across tasks.
Four tables with two policies and two tasks exhibit the conflict (App.~\ref{app:levels}). Much of an
evaluator's correlation can also be gained without changing a single decision. Adding a constant to
one task's scores leaves every pick, the regret, the MMRV and the order of the mean scores unchanged, while the pooled $r$ moves. Evaluators should be judged by the regret of the decision they serve.

\paragraph{Beyond the published tables.} The guarantees so far concern the published policies, tasks
and scores. Three extensions follow, each proved in the appendix.

\emph{Readers with fewer policies.} A reader rarely chooses among all $K$ policies: a lab may deploy
only some of them, and on some tasks only a few may apply. We call the policies available to a reader
its \emph{candidate subset}; for routing it may change from task to task. The evaluator picks its
top-scored policy within the subset, and regret is measured against the subset's best policy, which can
make the evaluator look much worse. On SIMPLER's visual-matching table, the evaluator loses
$1.7$ points per task when it chooses among all six policies, but $14.2$ with the worst two-policy
candidate subset on each task.  Call a pair \emph{violated} when one
policy has the higher outcome but not the strictly higher score. On each task, take the largest real
gap $\kappa_t$ of a violated pair. The \emph{worst rank violation} is the average
$\mathrm{WRV}=\frac1T\sum_t\kappa_t$. Its counterpart for deployment, $\mathrm{WRV}_{\mathrm{dep}}$, is the
largest gap between policy means over pairs violated by the mean scores.

\begin{theorem}[the worst candidate subset]\label{thm:menus}
Fix $K\ge2$, and let the evaluator pick its top policy within the candidate subset, ties broken against it.
The largest per-task regret over all candidate subsets is $\mathrm{WRV}$, and the largest deployment regret is $\mathrm{WRV}_{\mathrm{dep}}$; both are attained by subsets of two policies. Hence no bound that holds for every candidate subset, computed from any statistic, can be smaller.
\end{theorem}

The MMRV's bound $Kv$ holds for every candidate subset, and is at most $K$ times the least possible (App.~\ref{app:levels} relates WRV and MMRV).
$\mathrm{WRV}_{\mathrm{dep}}$ needs only the two vectors of policy means, which is one more reason
for item~2 of the card. On AutoEval's SIMPLER comparison, deployment costs nothing with all six
policies, and $36$ points with the worst candidate subset (App.~\ref{app:levels}).

\emph{New policies and new tasks.} Validations use a few policies and tasks, while a simulator is
used on new ones. Two results bound what any validation can claim beyond its tables
(App.~\ref{app:newtasks}). First, suppose policies are exchangeable. Then any $95\%$ bound on the regret of choosing among the $K$ published policies and a new one is too large to rule out any loss with probability at least
$1-.05(K+1)$, on every published table (Theorem~\ref{thm:newpolicy}). It can be informative every time only if $K\ge19$, the same threshold as split conformal prediction at $95\%$. Second, suppose tasks are drawn at random. Then
the bound of Theorem~\ref{thm:newtasks} carries a five-point guarantee over to new tasks, with
probability $95\%$, from $120$ tasks if the evaluator is perfect on the published table, and from
$333$ if its instance bound is two points (Cor.~\ref{cor:tasks}).  The robot
outcome tables of our corpus have at most $6$ policies and $17$ tasks, and those of the reported statistics at most
$8$ policies.

\emph{New runs of the evaluator.} A validation is meant to certify the evaluator, not one draw of
its scores, so the evaluator's rollouts, like the real trials, need confidence intervals. A simultaneous $95\%$ confidence box (Bonferroni over all cells) bounds the regret of the choice the evaluator makes when run to
convergence (Theorem~\ref{thm:joint}). AutoEval's autonomous scorer has no regret on the published
tables. Yet at its fifty trials per cell, the box guarantees only $30$ points. If its rates held, it
would need about a thousand trials per cell to guarantee five. The same box certifies that SIMPLER
loses at least $16.5$ points on one of AutoEval's tasks.

\paragraph{Using an evaluator before real tests.}To bound the regret
of the policy it deploys, a team can let the evaluator screen the top $k$ policies and test only those for real
(App.~\ref{app:practice}). If the published rates were exact, this would save from $21\%$ to $100\%$
of the hardware trials on seven of the eight robot validations, compared with testing every policy at
the same five-point standard with probability $.95$. It saves all of them when the evaluator alone
certifies the choice, and none on AutoEval's SIMPLER comparison. These savings are conditional: the hardware test accounts for its own trials, but the screen trusts the published rates, which at their trial counts cannot certify a choice within five points (Sec.~\ref{sec:exchange}).

%% file: discussion.tex
\section{Related work and limitations}\label{sec:related}

\paragraph{Validity and decisions.} Turning a validation statistic into a statement about a decision is
an old idea: for personnel selection under a bivariate normal model, \citet{taylorrussell1939} tabulated success ratios and \citet{brogden1949} showed that the gain is linear in the correlation. We take a worst case with no model. In robotics,
\citet{kadian2020sim2real} introduced the sim-versus-real correlation coefficient, the statistic whose guarantees we bound. The surrogate-endpoint and
proxy-reward literatures record that a high correlation with the target need not preserve the proxy's
decisions \citep{baker2003perfect,laidlaw2025correlated,liu2026robust}. For learned rewards, regret
bounds in the square root of one minus a correlation are known
\citep{gleave2021epic,skalse2024starc}. Ours hold on finite $K\times T$ tables on any bounded scale, with explicit constants tight to first order, an exact vacuity threshold, and an instance bound computed exactly.

Offline policy evaluation already reports regret at $k$ beside rank correlation
\citep{fu2021dope}, the kind of decision-level number we ask every validation to publish.

\paragraph{Judges and metrics.} Machine-translation meta-evaluation moved from correlation to pairwise
accuracy, then to a soft version that weighs each pair by its significance
\citep{mathur2020tangled,kocmi2021ship,deutsch2023ties,thompson2024spa}. Our results support that shift
in part: pairwise accuracy has a linear guarantee and orders WMT metrics by their regret about as well as the MMRV, but
it ignores the gaps between misordered policies, so its guarantee is looser. \citet{landesberg2026bestofn} observes that task difficulty inflates judges' pooled correlations; by Cor.~\ref{cor:vacuous}, a correlation no larger than the correlation ratio of the outcomes on tasks is vacuous on that table. Prediction-powered inference
corrects an evaluator's bias with paired labels \citep{angelopoulos2023ppi,badithela2025suresim}; it
estimates means, which a test after screening can use, but does not bound the regret of deciding by the
evaluator alone.

\paragraph{Limitations.}
\begin{itemize}[leftmargin=*,topsep=1pt,itemsep=1pt,parsep=0pt]
\item \emph{Scope.} Every guarantee applies to the published tables. It carries over to new tasks only
when they are drawn at random, which curated benchmarks usually are not, and then only with many
tasks. For a new exchangeable policy, a bound valid at level $\alpha$ can be informative on every table only if $K\ge 1/\alpha-1$ (Sec.~\ref{sec:q3},
App.~\ref{app:newtasks}), a worst case that WMT data do not approach (App.~\ref{app:wmtaudit}).
\item \emph{Noise.} Each guarantee inherits the noise of the published rates and their trial counts
(Sec.~\ref{sec:exchange}). The audits take their outcomes as exact.
\item \emph{Pessimism.} The worst case ranges over every score table with the published value,
including many no real evaluator would produce; Theorem~\ref{thm:noise} and Figure~\ref{fig:noise} show
how far below it an evaluator with random errors stays.
\item \emph{Evidence.} We never observe the decisions laboratories actually made; we bound their regret
from the published statistics. The corpus is small, with nine outcome tables, and on five of them one
policy is best on at least three quarters of the tasks, so choosing per task can gain little there.
\end{itemize}

\paragraph{Reproducibility.} The ancillary files of this arXiv submission hold the analysed tables with their sources and
licences, the criteria files, the library \texttt{evalceil} and every script; its \texttt{README.md} maps
each table and figure to its script. \texttt{regenerate.py} reruns every script offline and checks its outputs
against the released ones. Every instance bound except the NAS audit's is computed in exact rational arithmetic; value-only bounds, simulations, rank statistics and confidence boxes use floating point. The offline page \texttt{card.html} turns a reported correlation,
within-task average or MMRV, with $K$, into its value-only bound.

\paragraph{Conclusion.} A correlation says how closely an evaluator agrees with reality. It does not say
what a decision made with it can lose. A validation statistic is worth what it guarantees about the decision it
serves. A correlation guarantees little, and the better the evaluator, the more its guarantee
overstates the regret. With $K$ policies, it is vacuous unless $1-r^2$ is below $2.92/K$. For routing, the MMRV's guarantee is linear in its value; like any within-task statistic, it needs the policy means, or the tables, to certify deployment. Before trusting an evaluator for a decision,
convert its reported statistic into the regret it allows. If the validation reports only a
correlation, ask for the tables, or for the MMRV with each policy's real and evaluator means.

%% file: impact.tex
\subsubsection*{Broader Impact Statement}
This work re-analyses published validation evidence; it collects no new data and recruits no human participants. Its intended effect is safer use of evaluators: before acting on a choice made with a simulator, a metric or a judge, a laboratory can compute the largest regret that the evaluator's reported agreement allows. The main risk is misreading a guarantee: taking a worst case for an estimate, or a guarantee on the published tables for one on new tasks or policies; Sec.~\ref{sec:related} states this scope. The findings concern what published numbers can establish, not the conduct
of the authors who published them.

\subsubsection*{Use of AI Assistants}
Under the authors' direction, large language model assistants helped propose and refine hypotheses,
and contributed as follows.
\emph{Theory.} They proposed key ingredients of two results: for Theorem~\ref{thm:universal}, the
constant $\lambda^\ast$ of Lemma~\ref{lem:vaccol} and the step that uses Lemma~\ref{lem:rung} to reduce
the condition to one level $c$; for Theorem~\ref{thm:rankpooled}, the reduction of Step~3 of the proof
to the paired cells, a step they also wrote. Everything else in these proofs, and every other result and proof, is the
authors'.
\emph{Results used only by the code.} An assistant formulated and proved the four-level reduction used
by \texttt{evalceil/within.py}, and proposed the formulation, as a menu
of translated orthants, of the joint guarantee of a correlation and an MMRV published together
(\texttt{joint\_certificate}), whose proof is the authors'.
\emph{Experiments and code.} The four audits and their criteria files, the collection of reported
statistics, and the released library and tests were designed and carried out with substantial help from
assistants.
\emph{Writing.} Assistants edited the text, drafted parts of the appendices and captions, and suggested
how to relate the results to prior work; the authors revised every draft and decided what the paper claims.
\emph{Verification.} Assistants checked derivations independently, recomputed reported numbers and
compared citations with their sources. They generated none of the data: the language-model
judges and reward models among the evaluators are used as their authors released them.
The authors verified every result and are fully responsible for all content.

%% file: app_table.tex
\section{The guarantee on a published table}\label{app:table}

Every guarantee a correlation gives on a published table rests on two facts. Proposition~\ref{prop:cone} turns a
correlation into a squared distance to a cone, and Lemma~\ref{lem:rung} bounds that distance task by
task. This appendix proves both, then treats statistics that sum per-task terms (App.~\ref{app:separable}), deployment and top-$k$ screening (App.~\ref{app:coarse}), and why within-task orders leave a floor on the deployment regret (App.~\ref{app:levels}). $\Pi_{\mathcal X}$ is the projection onto a
closed convex set $\mathcal X$.  For each task, $g_t=p_{(1)t}-p_{(2)t}$ is the margin between its
two best policies, $L_t=p_{(1)t}-p_{(K)t}$ its range, $SS_t$ the centred sum of squares of its
column and $d_t(i)$ the squared distance from that column to $\mathcal C^t_i=\{u\in\mathbb R^K:u_i\ge u_j\ \forall j\}$, the score columns that rank $i$ first, so that
$d(a)=\sum_td_t(a(t))$.

\subsection{The cone and the per-task cost}\label{app:body}

\renewcommand*{\restatementname}{Proposition~\ref{prop:cone}}
\begin{restatement}[correlation and cone distance]
Let $K\ge2$, let $S>0$ and let $a$ be a choice.
Then $\max\{\mathrm{corr}(y,p):y\in\mathcal C_a,\ y\ \text{nonconstant}\}\le\sqrt{1-d(a)/S}$, the
maximum being attained.
If $d(a)<S$ this is an equality, attained at the projection $\Pi_{\mathcal C_a}(p)$.
If $d(a)=S$ and $T\ge2$, both sides are zero, the maximum being attained at a column-constant table; if $d(a)=S$ and $T=1$, the left side can be negative.
\end{restatement}

\begin{proof}[Proof of Proposition~\ref{prop:cone}]
Write $\tilde y,\tilde p$ for the centred tables.

\emph{The cone.} $\mathcal C_a$ is cut out by the linear homogeneous inequalities $y_{it}-y_{a(t)t}\le0$, each within one column, so it is a closed convex cone. Each inequality holds with equality on a column-constant matrix, so $\pm\mathbf 1\in\mathcal C_a$ and
$\mathcal C_a=\tilde{\mathcal C}\oplus\mathrm{span}(\mathbf 1)$ with
$\tilde{\mathcal C}=\mathcal C_a\cap\mathbf 1^{\perp}$. The component of $p$ along $\mathbf 1$ is matched exactly, so
$d(a)=\mathrm{dist}(\tilde p,\tilde{\mathcal C})^2$.

\emph{Reduction to a linear maximum.} $\mathrm{corr}(y,p)$ depends on $y$ only through the direction of $\tilde y$, and $y$ is nonconstant
exactly when $\tilde y\neq0$. So the maximum is
$\max\{\langle u,\tilde p\rangle/\sqrt S:u\in\tilde{\mathcal C},\lVert u\rVert_F=1\}$, attained
because that set is compact and, by $K\ge2$, nonempty.

\emph{The bound.} Let $q=\Pi_{\tilde{\mathcal C}}(\tilde p)$. By Moreau's decomposition \citep{moreau1962decomposition}, $\tilde p-q\perp q$, so $\lVert q\rVert_F^2=S-d(a)$, and $\langle\tilde p-q,u\rangle\le0$ on $\tilde{\mathcal C}$; with Cauchy-Schwarz, $\langle u,\tilde p\rangle\le\langle u,q\rangle\le\lVert q\rVert_F$ for unit $u\in\tilde{\mathcal C}$, with equality at $u=q/\lVert q\rVert_F$ when $q\neq0$.
So the maximum is $\sqrt{1-d(a)/S}$ when $q\neq0$, that is when $d(a)<S$, and at most zero
when $q=0$, since then $\langle u,\tilde p\rangle\le0$ on $\tilde{\mathcal C}$.

\emph{The case $d(a)=S$.} The column-mean table lies in every per-column cone, so
$d_t(i)\le SS_t$ and $d(a)\le\sum_tSS_t\le S$; equality forces $\bar p_{\cdot t}=\bar p$ for every $t$.
A column-constant $b$ with $b_{it}=\beta_t$ then has $\tilde b\in\tilde{\mathcal C}$ and
$\langle\tilde b,\tilde p\rangle=K\sum_t(\beta_t-\bar\beta)(\bar p_{\cdot t}-\bar p)=0$, so for $T\ge2$ a
nonconstant $\beta$ makes the maximum exactly zero.
For $T=1$ take $K=2$, $p=(1,0)^{\top}$ and $a(1)=2$, so $S=d(a)=\frac12$. Every $y\in\mathcal C_a$
has $\tilde y=(-s/2,s/2)$ with $s\ge0$, hence correlation $-1$ when nonconstant, while the right
side is zero.

\emph{Attainment without ties.} The maximiser $q$ ties each wrong pick with the best policy. If a choice $a$ attaining
$\varepsilon^\star(r)$ has $d(a)<S(1-r^2)$, its maximum correlation exceeds $r$, so a small perturbation of $q$ that
breaks every tie in favour of $a(t)$ keeps $a$ as the choice and the correlation above $r$. This fails only when
$S(1-r^2)$ equals the cost $d(a)$ of every choice attaining $\varepsilon^\star(r)$, which happens for finitely many $r$,
since there are finitely many choices. \end{proof}

\begin{lemma}[per-task option cost]\label{lem:rung} $d_t(i)\ge h_t(\ell)$ for every policy and
every column, where $\ell=p_{(1)t}-p_{it}$, $h_t(\ell)=\ell^2/2$ for $\ell\le2g_t$ and
$h_t(\ell)=\ell^2/2+(\ell-2g_t)^2/6$ above. The function $h_t$ is convex, increasing and never below $\ell^2/2$,
and it depends on the column only through $g_t$. When $\ell>0$, equality holds exactly when every entry
other than $p_{it}$, one leader and one runner-up is at or below the level
$c^{\star}$, which is $p_{(1)t}-\ell/2$ when $\ell\le2g_t$ and $p_{(1)t}-(\ell+g_t)/3$ above. If $i$ is a
runner-up, equality always holds: $d_t(i)=g_t^2/2$. \end{lemma}

\begin{proof}[Proof of Lemma~\ref{lem:rung}]
Fix a task $t$, write $v=p_{\cdot t}$ and $M=p_{(1)t}$, fix $i$ and put $\ell=M-v_i$, so that $c^{\star}$ of the statement is $M-\ell/2$ or
$M-(\ell+g_t)/3$.

\emph{Reduction to one variable.} Minimise $\lVert u-v\rVert^2$ over the cone by first fixing the level $c=u_i$: coordinate $i$
contributes $(c-v_i)^2$, and each $j\neq i$, constrained only by $u_j\le c$,
contributes $((v_j-c)_+)^2$. Hence
\begin{equation}\label{eq:phi}
  d_t(i)=\min_{c\in\mathbb R}\varphi_i(c),\qquad
  \varphi_i(c)=(c-v_i)^2+\sum_{j\neq i}\bigl((v_j-c)_+\bigr)^2 .
\end{equation}
Each summand is convex and the first coercive, so the minimum is attained at some $c^{\star}$, and
stationarity gives $c^{\star}-v_i=\sum_{j\neq i}(v_j-c^{\star})_+\ge0$.

\emph{The lower bound $h_t$.} For $\ell=0$ the bound is trivial; assume $\ell>0$. Every summand of
\eqref{eq:phi} is nonnegative, so keeping only a
leader $j^{\star}\ne i$ at $M$ and, when $\ell>g_t$, a runner-up $j^{\star\star}\notin\{i,
j^{\star}\}$ at $M-g_t$ leaves a lower bound on $d_t(i)$. Translating
$M$ to $0$, that bound is $\min_c(c+\ell)^2+((-c)_+)^2+((-g_t-c)_+)^2$, which is
\eqref{eq:phi} for the column $(0,-g_t,-\ell)$. (When $\ell\le g_t$, so $\ell=g_t$ and $i$ is a runner-up, only the leader's summand is kept; its minimum $\ell^2/2$ is the value found below for $\ell\le2g_t$.)
This minimand is convex and piecewise quadratic in $c$, with breakpoints at $0$ and $-g_t$.
 On $-g_t\le c\le0$ only the leader is active, and
$c=-\ell/2$ lies in the piece exactly when $\ell\le2g_t$, with value $\ell^2/2$. On
$c\le-g_t$ both are active, and $c=-(\ell+g_t)/3$
lies in the piece exactly when $\ell\ge2g_t$, with value
$\tfrac{2}{3}(\ell^2-\ell g_t+g_t^2)=\ell^2/2+(\ell-2g_t)^2/6$. The two cases cover
$[0,\infty)$ and agree at $\ell=2g_t$, giving $h_t$.

\emph{Shape of $h_t$.} $h_t'(\ell)$ is $\ell$ below $2g_t$ and $(4\ell-2g_t)/3$ above, both
nonnegative and equal to $2g_t$ at $\ell=2g_t$, and $h_t''\in\{1,4/3\}$. So $h_t$ is
increasing, convex and $C^{1}$, and exceeds $\ell^2/2$ exactly above $2g_t$.

\emph{Equality.} $c^{\star}$ is the unique minimiser of the bound that keeps only $i$, $j^{\star}$ and $j^{\star\star}$, and $\varphi_i$ lies above that bound. If every other entry is at or below $c^{\star}$, the two agree at $c^{\star}$, so $d_t(i)=h_t(\ell)$. An entry above $c^{\star}$ adds a positive term there, so $d_t(i)>h_t(\ell)$. On $\ell\le2g_t$ this is the
midpoint criterion $c^{\star}=(v_i+M)/2$, which the runner-up satisfies automatically, since
$-g_t\le-\ell/2$.

\emph{The runner-up identity.} Let $i$ be a runner-up. If $g_t=0$, $i$ is a
leader and both sides are zero. If $g_t>0$ the leader is unique and $\ell=g_t$ gives $h_t(g_t)=g_t^2/2$. Every
entry outside $\{i,j^{\star}\}$ is at most $v_i=p_{(2)t}<(v_i+M)/2=c^{\star}$, so the criterion
holds and $d_t(i)=g_t^2/2$.
\end{proof}

\subsection{Every statistic that sums per-task terms}\label{app:separable}

The MMRV and pairwise accuracy are sums of per-task terms, so each task's pick can be priced separately.
Ties in the evaluator's scores are broken against it. For task $t$ and policy $i$,
$\ell_t(i)=p_{(1)t}-p_{it}$ is the loss, $n_t(i)=\#\{j:p_{jt}>p_{it}\}$ the number of policies
strictly above $i$.

\paragraph{The frame.} An \emph{evaluator class} is a product of per-task sets: for each task $t$ and option $i$, a set $X_t(i)\subseteq\mathcal C^t_i$ of admissible score columns, $\mathcal C^t_i$ itself by default and $\{u:u_i>u_j\ \forall j\ne i\}$ for evaluators with a unique top policy. A score table is \emph{consistent} with a published value $v$ if its statistic is at least as good as $v$. The statistic \emph{splits over tasks} if these tables are exactly $\{y:\sum_t\varphi_t(y_{\cdot t})\le B\}$, or the same set with a strict inequality, for a budget $B$ and functions $\varphi_t:\mathbb R^K\to(-\infty,+\infty]$ that may depend on $p_{\cdot t}$. A sum of per-task gains $\sum_t\gamma_t\ge B'$ is the case $\varphi_t=-\gamma_t$. The \emph{per-option cost} is $c_t(i)=\inf\{\varphi_t(u):u\in X_t(i)\}$, \emph{attained} if the infimum is a minimum. The guarantee $G(B)$ is the largest regret $R(a)$ of a choice made by an admissible evaluator consistent with the value, its pick on each task being any of its top-scored policies.

\begin{proposition}[the exact guarantee of a statistic that splits over tasks]\label{prop:separable}
Let every $c_t(i)$ be finite, and let $\mathcal A(B)$ hold the choices with $\sum_tc_t(a(t))<B$ and, unless the published constraint is strict, the choices with $\sum_tc_t(a(t))=B$ whose every cost $c_t(a(t))$ is attained. Then (i) a choice is made by some admissible evaluator consistent with the value if and only if it is in $\mathcal A(B)$; (ii) $G(B)=\max\{R(a):a\in\mathcal A(B)\}$, a maximum over finitely many choices, attained by an admissible evaluator (the maximum of the empty set is $0$); (iii) with $\check G(B)$ and $\hat G(B)$ the same maximum over the choices with $\sum_tc_t(a(t))<B$ and with $\sum_tc_t(a(t))\le B$, $\check G(B)\le G(B)\le\hat G(B)$; these are the left and right limits of $G$ at $B$; $G(B)=\hat G(B)$ when every cost is attained and the constraint is not strict; and $G$ is a nondecreasing step function whose steps lie at the finitely many totals $\sum_tc_t(a(t))$.
\end{proposition}

\begin{proof}[Proof of Proposition~\ref{prop:separable}]
(i) If $\sum_tc_t(a(t))<B$, let $\delta>0$ be the difference and pick $u_t\in X_t(a(t))$ with $\varphi_t(u_t)<c_t(a(t))+\delta/T$; the table with these columns is admissible, consistent with the value, and makes the choice $a$. If the sum equals $B$ and every cost is attained, take minimisers. Conversely, an admissible $y$ consistent with the value that makes the choice $a$ has $y_{\cdot t}\in X_t(a(t))$, so $c_t(a(t))\le\varphi_t(y_{\cdot t})$ on every task and $\sum_tc_t(a(t))\le B$; at equality every one of these inequalities is tight, so every cost is attained, and a strict constraint excludes equality. (ii) Every consistent evaluator makes a choice in $\mathcal A(B)$ under any tie rule, and by (i) the maximising choice is made by one, its ties resolved against it. (iii) $\mathcal A(B)$ lies between the two sets of choices that define $\check G$ and $\hat G$. For $B'<B$ close enough to $B$ no total lies in $[B',B)$, so $\mathcal A(B')$ is the first set; for $B'>B$ close enough, it is the second. If every cost is attained and the constraint is not strict, $\mathcal A(B)$ is the second set. A larger budget enlarges $\mathcal A$, which changes only at a total.
\end{proof}

\paragraph{The pooled correlation.} The pooled correlation does not split over tasks, but it enters
through Prop.~\ref{prop:cone}: a choice is allowed exactly when $\sum_td_t(a(t))\le S(1-r^2)$, so its
per-option cost is $d_t(i)$ and its budget $S(1-r^2)$. This needs only an evaluator class that contains a
positive affine image of every score table, as scores confined to $[0,1]$ do.

\paragraph{The solver.} $\max\{R(a):\sum_tc_t(a(t))\le B\}$ is a multiple-choice knapsack \citep{kellerer2004knapsack}, one class per task and one item per policy. On a rational table with common denominator $D$, a dynamic program over the total loss solves it in $O(KT\Lambda)$ exact operations, $\Lambda\le1+D\sum_tL_t$. For each total it keeps the least cost and whether reaching it needs an unattained cost, and returns the largest total that Prop.~\ref{prop:separable}(i) admits. The released function \texttt{guarantee(stat, p, value)} takes the per-option cost as its only input specific to the statistic.

\paragraph{The MMRV.} The per-task costs $c^{\pm}_t(i)$ in the proof of Prop.~\ref{prop:rank}
(App.~\ref{app:mmrvproof}) are the MMRV's per-option costs, established and attained there, so its instance bound (Table~\ref{tab:eleven}) is Prop.~\ref{prop:separable} with budget $KTv$.

\paragraph{Pairwise accuracy.} Pairwise accuracy is the share of system pairs that the metric and the human scores order the same way \citep{kocmi2021ship,freitag2023wmt23metrics}; \citet{deutsch2023ties} add a variant that credits correctly predicted ties. As in the sign rule of \citet{kocmi2021ship}, a tie in the scores counts as a disagreement; unlike it, pairs tied in the outcomes are left out.

\begin{proposition}[pairwise accuracy]\label{prop:pairwise}
The per-option cost of pairwise accuracy, counted as the number of disagreeing pairs, is $c_t(i)=n_t(i)$, attained. Hence an accuracy of at least $A$ admits a per-task regret of at most $\min\{1,(1-A)K(K-1)/2\}$ on every outcome table, and this is a supremum.
\end{proposition}

\begin{proof}
Let $n=n_t(i)$. For $u\in\mathcal C^t_i$ and each of the $n$ policies $j$ with $p_{jt}>p_{it}$, $u_i\ge u_j$, so the pair disagrees; the column equal to $p_{\cdot t}$ except that entry $i$ is raised above every other entry disagrees on these pairs alone. So a wrong pick costs at least one, and with $N\le TK(K-1)/2$ pairs counted, an accuracy of at least $A$ allows at most $(1-A)N$ wrong picks, each losing at most $1$, so $R\le(1-A)K(K-1)/2$. For the supremum, let $f\le\min\{1,(1-A)K(K-1)/2\}$ with $fT$ an integer: on $fT$ tasks take the column $(1,x_2,\dots,x_K)$ with distinct $x_j$ tending to $0$ and pick the runner-up, and elsewhere the leader. Each such task has one disagreeing pair of $K(K-1)/2$, so the accuracy is $1-2f/(K(K-1))\ge A$ and the regret tends to $f$; letting $f$ increase to the cap over suitable $T$ approaches the bound.
\end{proof}

\paragraph{What the costs say.} A wrong pick that loses $\ell$ costs the correlation at least $\ell^2/2$ (Lemma~\ref{lem:rung}), so small mistakes are cheap. It costs the MMRV at least $\ell$, or $2\ell$ with a unique top policy, so its guarantee is linear. A pairwise statistic counts a number of pairs that does not depend on $\ell$, so the worst evaluator makes the largest mistakes; its guarantee is linear too, with constant $K(K-1)/2$ (Table~\ref{tab:rates}).

\paragraph{The MMRV's instance bound can be the looser.} On outcome columns $(1,.15,.25,0,.1)$ and $(.65,.85,1,.6,.7)$, an evaluator that ranks each leader first and permutes the rest has $r=.912$ and no regret, yet the MMRV charges every reordered pair: its instance bound, $.375$, exceeds the correlation's, $.200$ (\texttt{sims/extensions.py}).

\subsection{Deployment and top-\texorpdfstring{$k$}{k} screening}\label{app:coarse}

In this subsection an overline on a policy's row, as in $\bar y_i$, denotes its mean over tasks. Deployment and screening (Sec.~\ref{sec:setting}) depend on the score table only through the policy means. Split $y_{it}=\bar y_i+w_{it}$ with $\sum_t w_{it}=0$, and $p_{it}=\mu_i+q_{it}$ likewise. The cross terms vanish, so $\lVert y-p\rVert_F^2=T\lVert\bar y-\mu\rVert^2+\sum_{i,t}(w_{it}-q_{it})^2$, and any cone constraining $\bar y$ alone has squared distance $T$ times the squared distance from $\mu$ to the corresponding set in $\mathbb R^K$.

\paragraph{The aggregate and screen cones.} The aggregate cone $\{y:\bar y_i\ge\bar y_j\ \forall j\}$ is such a cone; Lemma~\ref{lem:rung}, applied to the policy means, gives its distance. So is the screening cone $\{y:\min_{i\in A}\bar y_i\ge\max_{j\notin A}\bar y_j\}$ of a set $A$ of $k$ policies. Both are closed and convex and contain every column-constant matrix, all that Prop.~\ref{prop:cone} uses. The screening set is the union over thresholds $\theta$ of the boxes $\{z\in\mathbb R^K:z_i\ge\theta$ on $A$, $z_j\le\theta$ off it$\}$, so its squared distance is $\min_\theta\bigl[\sum_{i\in A}((\theta-\mu_i)_+)^2+\sum_{j\notin A}((\mu_j-\theta)_+)^2\bigr]$, with $\theta$ in the role of $c$ in \eqref{eq:phi}. This is convex and piecewise quadratic in $\theta$, with breakpoints at the coordinates of $\mu$, so it is computed exactly in rational arithmetic; write $d(A)$ for $T$ times this squared distance.

\paragraph{The screen guarantee.} The regret of screening $A$ is $\max_j\bar p_j-\max_{i\in A}\bar p_i$, and the guarantee $\varepsilon^\star_k(r)$ is its maximum over the $k$-subsets with $d(A)\le S(1-r^2)$. At $k=1$ the two cones coincide. Table~\ref{tab:decisions} gives the guarantee at $k=2$ and the smallest $k$ meeting the tolerance.

\begin{corollary}[when a pooled correlation is vacuous]\label{cor:vacuous}
For $T\ge2$, $\varepsilon^\star(r)=\frac1T\sum_tL_t$, the worst choice's regret, whenever
\[
  r^2\;\le\;1-\tfrac1S\sum\nolimits_tSS_t\;=\;\tfrac KS\sum\nolimits_t(\bar p_{\cdot t}-\bar p)^2 ,
\]
the squared correlation ratio of $p$ on the tasks, the share of $S$ between the task means.
\end{corollary}

\begin{proof}
The column-mean vector lies in every per-column cone, so $d(a)\le\sum_tSS_t$ for every choice; once $S(1-r^2)$ is at least that sum, \eqref{eq:exchange} excludes no choice. The equality is the one-way analysis-of-variance decomposition $S=\sum_tSS_t+K\sum_t(\bar p_{\cdot t}-\bar p)^2$.
\end{proof}

The condition is sufficient, not necessary:
AutoEval's SIMPLER validation has a correlation above it, $r=.450$, and still has the guarantee $\frac1T\sum_tL_t$. Across the corpus, this share runs from $.149$ to $.868$. The only outcome table where the condition explains such a guarantee is AgentRewardBench, with the highest share and a threshold of
$.9318$: its rule-based checker's $.9314$ is just below it, its judge's $.9648$ above. The condition involves the outcome table alone, so a benchmark can check it before scoring any evaluator.

\subsection{Deployment: levels against orders}\label{app:levels}

A statistic that sees only the order within each task, as the MMRV, every pairwise statistic and the task-averaged Kendall and Spearman coefficients do, cannot guarantee a deployment regret below $\Phi(p)$. The pooled correlation sees the levels across tasks.

\begin{proof}[Proof of Theorem~\ref{thm:levels}]
Let $U$ hold the Pareto-optimal policies and let $\mu_i$ be policy $i$'s mean outcome, so that $\Phi(p)=\max_j\mu_j-\min_{i\in U}\mu_i$. For $i\in U$ let $y_{kt}=p_{kt}+(T+2)\mathbf 1[p_{kt}\ge p_{it}]$. Each column of $y$ is a strictly increasing function of the same column of $p$, so every within-task weak order, and with them the statistic, is that of $p$. Row $i$ gains the step $T+2$ on every task. A policy with another row of $p$ does not dominate $i$, so it lies strictly below $i$ on some task and gains at most $T-1$ steps; its outcomes, in $[0,1]$, recover at most $T$ of the deficit. A policy with the same row ties with $i$. So the evaluator deploys a policy of mean $\mu_i$.
\end{proof}

$\Phi$ is a lower bound on the guarantee, not its value. Some statistics do not penalise an evaluator for reordering policies the outcome table ties: the MMRV, whose term for a tie is its zero gap, and pairwise accuracy with \citet{kocmi2021ship}'s sign rule, when tied pairs are left out. Such a statistic leaves $\Phi_0(p)=\max_j\mu_j-\min_{i\in U_0}\mu_i\ge\Phi(p)$ open, where $U_0$ holds the weakly Pareto-optimal policies, those that no other policy beats on every task. At an MMRV of $0$ or an accuracy of $1$ this is the exact guarantee: a policy that another beats on every task scores below it on every task, so it is never deployed. The construction above with $y_{kt}=p_{kt}+(T+2)\mathbf 1[p_{kt}>p_{it}\text{ or }k=i]$ breaks every tie in favour of $i$ at an MMRV of $0$ and an accuracy of $1$. The within-task average of correlations (columns nonconstant) is invariant under a positive scale per task. With $y_{\cdot t}=\alpha_tp_{\cdot t}$, $\alpha_t>0$, the same argument replaces $U$ by the supported Pareto-optimal policies, which maximise some positive weighting of the tasks. Table~\ref{tab:sepship} gives $\Phi$ and $\Phi_0$; on AutoEval's autonomous table, for instance, $\Phi=.112$ and $\Phi_0=.408$.

\begin{proof}[Proof of Proposition~\ref{prop:noscalar}]
Take $T=2$ and first $K=2$, policies as rows. Table $X$ has $p=\bigl(\begin{smallmatrix}.6&0\\0&.5\end{smallmatrix}\bigr)$ and $y=\bigl(\begin{smallmatrix}.1&0\\0&.9\end{smallmatrix}\bigr)$: both tasks are ordered correctly, so the per-task regret is $0$, while the mean scores deploy the second policy, of mean $.25$ against $.30$, a deployment regret of $.05$. Table $Z$ has $p=\bigl(\begin{smallmatrix}1&1\\0&0\end{smallmatrix}\bigr)$ and $y=\bigl(\begin{smallmatrix}0&1\\.1&0\end{smallmatrix}\bigr)$: the evaluator picks the second policy on the first task, a per-task regret of $\frac12$, while its means $(.5,.05)$ deploy the first, a deployment regret of $0$. Tables $U$ and $V$ have the same $p$ as $Z$, and $y=p$ or $y=\bigl(\begin{smallmatrix}0&0\\1&1\end{smallmatrix}\bigr)$: both regrets are $0$ on $U$ and $1$ on $V$. If $G$ is nondecreasing and $G(s)$ is strictly smaller on one pair than on another, so is $s$; if $G$ is nonincreasing, $s$ is strictly larger. If $G_1$ and $G_2$ are monotone in the same direction, $X$ and $Z$ force opposite orders of $s(X)$ and $s(Z)$; if in opposite directions, $U$ and $V$ force opposite orders of $s(U)$ and $s(V)$. For $K>2$ add policies with outcome $0$ on every task and scores below all others; no regret changes.
\end{proof}

\renewcommand*{\restatementname}{Theorem~\ref{thm:menus}}
\begin{restatement}[the worst candidate subset, in full]
\emph{(i)} As stated in Sec.~\ref{sec:q3}. \emph{(ii)}
$\mathrm{MMRV}\le\mathrm{WRV}\le K\,\mathrm{MMRV}$, and $\mathrm{WRV}\le\frac K2\mathrm{MMRV}$ when every
task with a violated pair has a largest violated pair that the scores order strictly the wrong way.
\emph{(iii)} No single statistic gives both $\mathrm{WRV}$ and $\mathrm{WRV}_{\mathrm{dep}}$ through
monotone maps.
\end{restatement}

\begin{proof}[Proof of Theorem~\ref{thm:menus}]
\emph{(i)} Fix a task and a candidate subset $M$. Let $a$ be the
evaluator's pick in $M$ and $b$ a policy of $M$ with the highest outcome. If $p_{bt}=p_{at}$ the regret is
$0$. Otherwise $p_{bt}>p_{at}$, and since $a$ has a top score in $M$, $y_{bt}\le y_{at}$, so $(b,a)$ is
violated and the regret $p_{bt}-p_{at}$ is at most $\kappa_t$. If $\kappa_t>0$, let $(i,j)$ be a violated
pair with gap $\kappa_t$ and take the candidate subset $\{i,j\}$. Then $y_{it}\le y_{jt}$, and the evaluator picks $j$:
by its score if $y_{it}<y_{jt}$, and by the tie rule otherwise. Its regret is $\kappa_t$. Candidate subsets are chosen
task by task, so the largest mean regret is the mean of the largest, $\mathrm{WRV}$. The same argument on
the policy means, with one candidate subset, gives $\mathrm{WRV}_{\mathrm{dep}}$. A bound that holds for every candidate subset
holds in particular for the worst one, which proves the remaining claim.

\emph{(ii)} A term of the MMRV on task $t$ is a gap $\lvert p_{it}-p_{jt}\rvert$ with
$[y_{it}>y_{jt}]\ne[p_{it}>p_{jt}]$. If the gap is positive, then $(i,j)$ or $(j,i)$ is violated: if
$p_{it}>p_{jt}$ then $y_{it}\le y_{jt}$, and if $p_{jt}>p_{it}$ then $y_{it}>y_{jt}$. So the term is at most $\kappa_t$. Every one of the $K$ terms is at most
$\kappa_t$, so $\mathrm{MMRV}\le\mathrm{WRV}$. Conversely, for a violated pair $(i,j)$ with gap $\kappa_t$
we have $[y_{it}>y_{jt}]=0\ne1=[p_{it}>p_{jt}]$, so $i$'s term is at least $\kappa_t$. If
$y_{it}<y_{jt}$, $j$'s term is also at least $\kappa_t$. Summing the $K$ terms and averaging over tasks
gives $\mathrm{WRV}\le K\,\mathrm{MMRV}$, and $\frac K2\mathrm{MMRV}$ when on every such task some such pair has
$y_{it}<y_{jt}$.

\emph{(iii)} With two policies a candidate subset is the full set or one policy, and one policy costs nothing. So
$\mathrm{WRV}$ and $\mathrm{WRV}_{\mathrm{dep}}$ are the realised regrets, and the four tables in the
proof of Proposition~\ref{prop:noscalar} apply. The padding policies, scored below all others and with
outcome $0$, form no violated pair with any policy, task by task or in the means, so the
argument holds for every $K$. A brute-force check over every candidate subset, on $3{,}000$ random tables with up to
four policies and three tasks and with forced ties, confirms (i) and (ii) (\texttt{sims/menus.py}).
\end{proof}

\emph{On the corpus.} The worst candidate subset costs more than the full one per task on eight of the eleven
validations, and for deployment on four: SIMPLER's visual matching ($0$ against $7.2$ points), its variant
aggregation ($8.7$ against $20.7$), AutoEval's SIMPLER comparison ($0$ against $36$) and WMT23 ($.7$
against $2.1$). The bound $\frac K2v$ covers the full candidate subset only: on WorldGym it is $.072$, below the
worst candidate subset's $.091$, because on four tasks the only largest violated pair is tied in scores.

%% file: app_worst.tex
\section{Worst cases from a published statistic alone}\label{app:worst}

\subsection{Proof of Theorem~\ref{thm:everyboard}: worst cases over all outcome tables}\label{app:universal}

This subsection proves Theorem~\ref{thm:everyboard}. Theorem~\ref{thm:universal} treats the pooled correlation, giving parts (ii) and (iii) with explicit constants and the threshold $\rho_1(K)$ behind $r_{\mathrm{vac}}$. Theorem~\ref{thm:rankpooled} gives part (iv) (pooled rank correlations) and Corollary~\ref{cor:mmrvexact} part (i) (MMRV); a final proof assembles them. Table~\ref{tab:needed} gives the correlation needed for each $K$ and tolerance. We start from the simplest bound, which Theorem~\ref{thm:everyboard} sharpens.

\begin{corollary}[a regret bound from the correlation and the policy count alone]\label{cor:boardfree}
For every $K\ge2$, every $T$ and every outcome table $p\in[0,1]^{K\times T}$,
$\varepsilon^\star(r)\le\sqrt{K(1-r^2)/2}$, a bound that is at least $1$ exactly when $r\le\sqrt{1-2/K}$.
\end{corollary}

\begin{proof}
Let $a$ be a choice with $d(a)\le S(1-r^2)$ (Prop.~\ref{prop:cone}). Lemma~\ref{lem:rung} and
Cauchy-Schwarz give $d(a)\ge\frac12\sum_t\ell_t(a(t))^2\ge\frac T2R(a)^2$, and $S\le KT/4$ by Popoviciu's inequality. Hence $R(a)^2\le\tfrac{K}{2}(1-r^2)$, at least $1$ exactly when
$K(1-r^2)\ge2$.
\end{proof}

\input{table_needed}

\paragraph{Columns.} A \emph{column} is a vector $v\in[0,1]^K$ (not the MMRV value $v$) with a \emph{chosen entry} $i$.
We write $x=v_i$ for the chosen value and $v_{\max}=\max_kv_k$ for the leader's value.
The loss is $\ell=v_{\max}-x$. The quantity $\Psi=\sum_kv_k(1-v_k)$
vanishes exactly on $0/1$ columns. Let $s_K=\sqrt{K\rho/2}$ (Cor.~\ref{cor:boardfree}) and
$V_K(\rho)=\sup\{\varepsilon^\star_p(r):T\ge1,\ p\in[0,1]^{K\times T}\}$, with $\varepsilon^\star_p$
the guarantee \eqref{eq:exchange} on $p$; by Prop.~\ref{prop:cone}, $V_K$ is also the supremum of
$R(a)$ over all outcome tables and score tables with $\corr(y,p)\ge r$.

\paragraph{Reduction to laws over columns.} The worst case over tables equals the worst case over
distributions of columns, which are easier to optimise over. A \emph{law} is a distribution over pairs (column,
chosen entry). A table and a choice give the empirical law $\nu$ of $(p_{\cdot t},a(t))$, with
$R(a)=\E_\nu\ell$, $d(a)/T=\E_\nu d$ and $S/T=\mathrm{Var}(\nu)=\min_\lambda\E_\nu\sum_k(v_k-\lambda)^2$.
So every choice the correlation allows gives a law with $\E_\nu d\le\rho\,\mathrm{Var}(\nu)$, \emph{the
constraint}. Conversely, a finitely supported law with rational weights meeting the constraint
strictly is the empirical law of a table (repeat each column in proportion to its weight). A law
meeting it with equality is a limit of such tables: mix in, with weight $\eta$, a zero-loss column of
positive variance (cone cost zero; $\mathrm{Var}$ is concave in $\nu$, so the constraint becomes
strict), then round the weights to nearby rationals, which keeps it strict by continuity. The regret
tends to $\E_\nu\ell$ as $\eta\to0$. So $V_K(\rho)$ is the supremum of $\E_\nu\ell$ over finitely
supported laws meeting the constraint.

\begin{theorem}[what a correlation guarantees on every outcome table]\label{thm:universal}
Let $K\ge2$, $\rho\in(0,1)$,
$\beta_K(\rho)=\dfrac{K\rho/2}{\rho+\sqrt{\rho^2+(1-2\rho)K\rho/2}}$, $B_K=\min\{1,\beta_K\}$ for
$\rho\le\frac12$ and $B_K=\min\{1,K\rho/2\}$ above, $\underline B_K=2s_K/(2r+s_K)$ and
$\rho_1(K)=\bigl(2(\sqrt{2K-1}-\sqrt{K-1})^2\bigr)^{-1}$.
\emph{(i)} On every outcome table $\varepsilon^\star_p(r)\le V_K(\rho)\le B_K(\rho)$, and $B_K<s_K$
whenever $s_K<1$.
\emph{(ii)} $V_2=B_2$ on $(0,1)$, and $V_3=B_3$ on $(0,\frac14]$.
\emph{(iii)} For $K\ge4$, $V_K\ge\min\{1,\underline B_K\}$. Consequently, for every $K$,
$s_K-\max\{\tfrac K4,1\}\rho+O(\rho^{3/2})\le V_K(\rho)\le s_K-\rho+O(\rho^{3/2})$ as $\rho\to0$ with $K$ fixed. Hence the bound $s_K$ of
Cor.~\ref{cor:boardfree} is exact to first order.
\emph{(iv)} For $K\ge3$, $V_K(\rho)=1$ if and only if $\rho\ge\rho_1(K)$, and $\rho_1(K)>2/K$.
For $K=2$, $V_2<1$ on $(0,1)$.
\emph{(v)} For $K\ge3$, $\rho_1(K)=\bigl(\sqrt{2K-1}+\sqrt{K-1}\bigr)^2/(2K^2)$ and
$C/(2K)-C'/(2K^2)\le\rho_1(K)<C/(2K)$, with $C=3+2\sqrt2$ and $C'=2+3\sqrt2$. So for $r\in(0,1)$ and
$K\ge3$: if $K\ge C/(2(1-r^2))$ the correlation is vacuous, and if $C/(2K)-C'/(2K^2)>1-r^2$ it
guarantees a regret below one.
\end{theorem}

By part (iv), $r_{\mathrm{vac}}(K)=\sqrt{1-\rho_1(K)}$. This is below $\sqrt{1-2/K}$, where the bound of Cor.~\ref{cor:boardfree} becomes vacuous, so a correlation between the two still guarantees a regret below one.
Part (v) gives the scaling in $K$: $1-r_{\mathrm{vac}}(K)^2$ falls like $C/(2K)$, just under $2.92/K$. So a correlation $r$ guarantees anything about routing only for fewer than $C/(2(1-r^2))$
policies. By part (iv) the exact limit is fourteen policies at $r=.9$ and twenty-nine at $r=.95$.
Part (iv) rests on a lower bound for the cone cost of columns that lose everything.

\begin{lemma}[maximal-loss columns]\label{lem:vaccol}
Let $K\ge3$ and $\lambda^\ast=\sqrt{K-1}\big/\bigl(\sqrt{K-1}+\sqrt{2K-1}\bigr)$. Every column with $v_{\max}=1$
and $x=0$ has $d\ge\rho_1(K)\sum_k(v_k-\lambda^\ast)^2$.
\end{lemma}

\begin{proof}[Proof of Lemma~\ref{lem:vaccol}]
Write $\rho=\rho_1(K)$ and $\lambda=\lambda^\ast$. By Lemma~\ref{lem:rung}, the projection sets the chosen
entry and every entry above some level $c\in[0,1]$ to $c$. For fixed $c$, the $K-2$ other entries
$f$ contribute separately, so it suffices that
$G(c)=c^2+(1-c)^2-\rho(\lambda^2+(1-\lambda)^2)+(K-2)\phi(c)\ge0$ on $[0,1]$, where
$\phi(c)=\min_{f\in[0,1]}[(f-c)_+^2-\rho(f-\lambda)^2]$. The bracket is concave on $[0,c]$ (minimum at
$f=0$ or $f=c$) and convex on $[c,1]$ with stationary point $f^\ast=(c-\rho \lambda)/(1-\rho)$. So $\phi$
is the least of three branches, at $f=0$, at $f=f^\ast$ (valid for $\lambda \le c\le1-\rho(1-\lambda)$)
and at $f=1$; the value at $f=c$ is never smaller than all three. Write $G_0,G_\ast,G_1$ for $G$ on each branch. $G_0$ is convex with minimum at $c=\frac12$, equal to
$\tfrac12-\rho\bigl((K-1)\lambda^2+(1-\lambda)^2\bigr)=0$ by $(K-1)^2+(2K-1)=K^2$ and $(2K-1)-(K-1)=K$. $G_1$ is convex
with minimum zero at $c=(K-1)/K$, since $(2K-1)+1=2K$. $G_\ast$ has leading coefficient
$(K\rho-2)/(\rho-1)\le0$ since $K\rho\ge2$, so it is concave and we check its ends:
$(1-\rho)(\lambda^2+(1-\lambda)^2)>0$ at $c=\lambda$, and $G_\ast=G_1\ge0$ at the other, where $f^\ast=1$. It remains to show that $2/K<\rho<1$. As
$\sqrt{2K-1}-\sqrt{K-1}=K\big/\bigl(\sqrt{2K-1}+\sqrt{K-1}\bigr)$ increases with $K$,
$\rho=1\big/\bigl(2(\sqrt{2K-1}-\sqrt{K-1})^2\bigr)$ decreases, so $\rho\le\rho_1(3)<.75$. And
$\rho>2/K$ is equivalent to $8\sqrt{(K-1)(2K-1)}>11K-8$, which holds as $64(K-1)(2K-1)-(11K-8)^2=K(7K-16)>0$ for
$K\ge3$.
\end{proof}

\begin{proof}[Proof of Theorem~\ref{thm:universal}]
\emph{Part (i).} A large loss requires a large variance in the column, and the constraint caps that variance. The first inequality is the definition of $V_K$. Every column has
$\Psi\ge\ell(1-\ell)$: for $\ell>0$, the terms of the leader and the chosen entry give
$\Psi\ge v_{\max}(1-v_{\max})+(v_{\max}-\ell)(1-v_{\max}+\ell)$, concave in $v_{\max}\in[\ell,1]$ and equal to $\ell(1-\ell)$ at both
ends. For a law with mean entry $\lambda$, $\mathrm{Var}(\nu)=K\lambda(1-\lambda)-\E \Psi\le\tfrac K4-\E\ell+\E\ell^2$.
With $d\ge\ell^2/2$ (Lemma~\ref{lem:rung}), $R=\E\ell$ and $m_2=\E\ell^2$, the constraint gives
$(\tfrac12-\rho)m_2+\rho R\le K\rho/4$. If $\rho\le\frac12$,
$q_\rho(\ell)=(\tfrac12-\rho)\ell^2+\rho\ell$ is increasing and convex, so Jensen gives
$q_\rho(R)\le K\rho/4$, that is $R\le\beta_K$, the positive root of $q_\rho=K\rho/4$. If
$\rho>\frac12$, the coefficient of $m_2$ is negative and $m_2\le R$, so $R\le K\rho/2$. Regret is
at most one. For $s_K<1$ and $\rho\le\frac12$, $\beta_K<s_K$ is equivalent to
$s_K-\rho<\sqrt{\rho^2+(1-2\rho)s_K^2}$, which holds if $s_K\le\rho$ and otherwise, after squaring,
is equivalent to $s_K<1$. For $\rho>\frac12$, $K\rho/2=s_K^2<s_K$.

\emph{Part (ii).} We build laws on which every inequality of part (i) is an equality.
\emph{Case $K=2$, $\rho\le\frac12$.} Let $\ell=\beta_2$ and take the columns $(1,1-\ell)$ and $(\ell,0)$, the second
policy chosen on both. Each has cone cost $\ell^2/2$ (both entries move to their midpoint), so
$d(a)=\ell^2$, while $S=1-2\ell+2\ell^2$; $q_\rho(\ell)=\rho/2$ is exactly $d(a)=\rho S$, and the
regret is $\ell$.
\emph{Case $K=2$, $\rho>\frac12$.} Put weight $\rho$ on $(1,0)$ choosing the zero and
$\tfrac{1-\rho}2$ on each of $(1,1)$ and $(0,0)$; the regret is $\rho=K\rho/2$ with equality in the
constraint.
\emph{Case $K=3$, $\rho\le\frac14$.} Let $\ell=\beta_3\le\frac12$, and put weight
$w=(\tfrac32-\ell)/(2-2\ell)\le1$ on $(1,1-\ell,0)$ and $1-w$ on $(\ell,0,0)$, the second entry
chosen. Then $\lambda=\frac12$, $\Psi=\ell(1-\ell)$ and $d=\ell^2/2$ on both columns, so the law attains
$\beta_3$. By the reduction, each law is a limit of outcome tables.

\emph{Part (iii).} We build a law on columns with $0$ or $K-2$ middle entries $h$ that reaches $\underline B_K$. Let $x=1-\ell$, $h=1-\ell/2$ and $v_j=(1,x,h,\dots,h,0,\dots,0)$ with $j$
entries $h$, the entry $x$ chosen. The closest column ranking $x$ first moves $x$ and the leader to
$h$ and nothing else, so $d=\ell^2/2$ for every $j$. Mix $j=0$ with weight $w=K/(2(K-2))$ ($\le1$
exactly when $K\ge4$) and $j=K-2$ with weight $1-w$. This law has mean column sum $Kh/2$ and
$\mathrm{Var}=\ell^2/2+\tfrac K4(1-\ell/2)^2$, and $d/\mathrm{Var}=\rho$ solves to $\ell=\underline B_K$. The
ratio $d/\mathrm{Var}=1/\bigl(1+\tfrac K2(1/\ell-1/2)^2\bigr)$ increases with $\ell$. If $\underline B_K\le1$
the law at $\ell=\underline B_K$ meets the constraint with regret $\underline B_K$; if $\underline B_K>1$ the law at $\ell=1$ has a
smaller ratio and regret $1$. So $V_K\ge\min\{1,\underline B_K\}$. For the expansions,
$r=1-\rho/2+O(\rho^2)$ gives $\underline B_K=s_K-s_K^2/2+O(s_K^3)=s_K-\tfrac K4\rho+O(\rho^{3/2})$, and $\beta_K=s_K-\rho+O(\rho^{3/2})$.
For $K\le3$, part (ii) gives $V_K=\beta_K$ near zero, so the lower estimate holds with constant $1$.

\emph{Part (iv), sufficiency.} A mixture of two $0/1$ columns with full loss meets the constraint once $\rho\ge\rho_1(K)$. Let $u_j$ be the $0/1$ column with $j$ ones, the chosen entry a zero
and a one the leader, so $\ell=1$ and the projection moves the chosen zero and the $j$ ones
to $j/(j+1)$, so $d=j/(j+1)$. Mixing $u_{K-1}$ (weight $w$) and $u_1$ gives regret one and meets the constraint if and only if $\E d/\mathrm{Var}\le\rho$. With $z=K+w(K-2)$, the ratio is
$z/\bigl(2(z-K+1)(2K-1-z)\bigr)$, minimised at $z=\sqrt{(K-1)(2K-1)}$,
which lies in $[K,2K-2]$ (the range of $z$) for $K\ge3$; the minimum is $\rho_1(K)$. The reduction
then gives tables with regret tending to one.

\emph{Part (iv), necessity.} Below $\rho_1$, no allowed law can put much weight near full loss. Let $K\ge3$, $\rho<\rho_1$ and $\lambda^\ast$ as in Lemma~\ref{lem:vaccol}. A
\emph{maximal-loss} column has $v_{\max}=1$ and $x=0$; by Lemma~\ref{lem:vaccol} it has
$d\ge\rho_1\sum_k(v_k-\lambda^\ast)^2$. Put $\psi(v,i)=d(v,i)-\rho\sum_k(v_k-\lambda^\ast)^2$. A law meeting
the constraint has $\E\psi\le\E d-\rho\,\mathrm{Var}\le0$, as $\mathrm{Var}$ is a minimum over
$\lambda$. On a maximal-loss column $\sum_k(v_k-\lambda^\ast)^2\ge(1-\lambda^\ast)^2+\lambda^{\ast2}\ge\frac12$, so
$\psi\ge\kappa=(\rho_1-\rho)/2>0$. Fix a small constant $\delta>0$. If $\ell\ge1-\delta$, setting the leader to one and the chosen entry to zero moves the column by at most
$\sqrt2\delta$ to a maximal-loss column; $\sqrt d$ is $1$-Lipschitz and at most $\sqrt K$ on the cube,
so both terms of $\psi$ are $2\sqrt K$-Lipschitz, and $\psi\ge\kappa/2$ for $\delta$ small enough.
Everywhere $\psi\ge-\rho K$. Splitting $\E\psi\le0$ on $\{\ell\ge1-\delta\}$ gives
$\tfrac\kappa2\Pr[\ell\ge1-\delta]\le\rho K(1-\Pr[\ell\ge1-\delta])$, so the probability is at most
$\pi_0=\rho K/(\rho K+\kappa/2)<1$ for every allowed law, and $\E\ell\le1-\delta(1-\pi_0)<1$
uniformly, so $V_K<1$. The proof of Lemma~\ref{lem:vaccol} shows that $\rho_1>2/K$. For $K=2$,
part (ii) gives $V_2=B_2<1$.

\emph{Part (v).} Write $m=\sqrt{2K-1}$ and $n=\sqrt{K-1}$. Since $m-n=(m^2-n^2)/(m+n)=K/(m+n)$,
$\rho_1=1/(2(m-n)^2)=(m+n)^2/(2K^2)$, and $(m+n)^2=3K-2+2\sqrt{2K^2-3K+1}$. As $2K^2-3K+1<2K^2$,
$(m+n)^2<(3+2\sqrt2)K$, the upper bound. For the lower bound, $\sqrt{x^2-y}\ge x-y/x$ whenever
$0\le y\le x^2$ (square both sides if the right side is positive); with $x=\sqrt2K$ and $y=3K-1$,
$\sqrt{2K^2-3K+1}\ge\sqrt2K-(3K-1)/(\sqrt2K)\ge\sqrt2K-3/\sqrt2$, so
$(m+n)^2\ge(3+2\sqrt2)K-(2+3\sqrt2)$. Dividing by $2K^2$ gives both bounds. By part (iv) the correlation is vacuous exactly when $1-r^2\ge\rho_1(K)$. If $K\ge C/(2(1-r^2))$, then $1-r^2\ge C/(2K)>\rho_1(K)$. If instead $1-r^2<C/(2K)-C'/(2K^2)\le\rho_1(K)$, part (iv) gives $V_K<1$.
\end{proof}

\paragraph{Pooled rank correlations.} Kendall's $\tau$ over $N$ tie-free cells is one minus twice the share of
\emph{discordant} pairs (cells the scores order opposite to the outcome table); Spearman's
$r_{\mathrm S}$ is the Pearson correlation of the two rankings.

\begin{theorem}[pooled rank correlations]\label{thm:rankpooled}
Let the outcome table and the scores be free of ties, with $N=KT$ cells, and let $\tau$ and $r_{\mathrm S}$ be
Kendall's and Spearman's coefficients over all cells. Every taskwise maximiser $a$ has
\emph{(i)} $R(a)\le K\sqrt{(1-\tau)/2}$, and \emph{(ii)} $R(a)<K\bigl((1-r_{\mathrm S})/4\bigr)^{1/3}$
when $R(a)>0$. Both are tight as $T\to\infty$: the supremum of $R$ over outcome and score tables with Kendall
coefficient at least $\tau$ is $\min\{1,K\sqrt{(1-\tau)/2}\}$, and with Spearman coefficient at least
$r_{\mathrm S}$ it is $\min\{1,K((1-r_{\mathrm S})/4)^{1/3}\}$. So the statistic is vacuous (worst case $1$) exactly when
$1-\tau\ge2/K^2$, respectively $1-r_{\mathrm S}\ge4/K^3$.
\end{theorem}

\begin{proof}[Proof of Theorem~\ref{thm:rankpooled}]
\emph{Step 1.} The regret is at most the largest share of tasks that straddle one threshold.
Let $M(\theta)$ count the tasks whose chosen cell lies below $\theta$ and whose leader lies at or
above it. On task $t$ with leader $b$, $\ell_t=\int_0^1\mathbf 1[p_{a(t)t}<\theta\le
p_{bt}]\,d\theta$, so $R(a)$ is the average of $M(\theta)/T$ over $\theta$, and
$R(a)\le\sup_\theta M(\theta)/T$. Each such task gives a \emph{straddling pair}
(chosen cell, leader); these pairs are disjoint, and in each the chosen cell is scored above the
leader.

\emph{Step 2 (Kendall).} Many straddling pairs force many discordant pairs. If $M$ disjoint pairs
$(u_i,w_i)$ straddle a common $\theta$ with $y(u_i)>y(w_i)$, there are at least $M(M+1)/2$
discordant pairs. Each $(u_i,w_i)$ is discordant. A cell $c$ with $p(u_i)<p(c)<p(w_i)$ is
discordant with $u_i$ if $y(c)<y(u_i)$ and with $w_i$ otherwise. Let $\mathrm{gap}_i$ be the
truth-rank distance between $u_i$ and $w_i$. The $u$'s occupy distinct truth ranks below $\theta$
and the $w$'s distinct ranks above, so $\sum_i(\mathrm{gap}_i-1)\ge M^2-M$. Each discordant pair
beyond the $M$ straddling ones is counted at most twice in this sum, so the count is at least
$M+\frac12(M^2-M)$.

\emph{Step 3 (Spearman).} The same pairs force a large sum of squared rank differences. Under the same assumption, let $\Delta_x$ be the score rank minus the truth
rank of cell $x$. Then $\sum_x\Delta_x^2\ge M(M+1)(2M+1)/3$. Dropping the other cells
and re-ranking within the $2M$ paired cells does not increase $\sum \Delta_x^2$ (each full rank is the rank within the paired cells plus the count of outside cells below it, which is nondecreasing in that rank; the cross term is then nonnegative by the rearrangement inequality). Then the $u$'s have truth ranks $1,\dots,M$ and the $w$'s
$M+1,\dots,2M$. Sort the score ranks $s_1<\dots<s_M$ of the $u$'s and $s'_1<\dots<s'_M$ of the $w$'s.
The $M-j+1$ $w$'s with score rank at least $s'_j$ have distinct partners scored higher, so
$s_j>s'_j$; hence $s_j$ exceeds $s_1,\dots,s_{j-1},s'_1,\dots,s'_j$, so $s_j\ge2j$, and likewise
$s'_j\le2j-1$. The monotone matching minimises each side's sum of squares, so
$\sum\Delta^2\ge\sum_j(s_j-j)^2+\sum_j(M+j-s'_j)^2\ge\sum_jj^2+\sum_j(M+1-j)^2$.

\emph{Part (i).} There are $(1-\tau)N(N-1)/4$ discordant pairs, so at the maximising $\theta$
Step 2 gives $M(M+1)/2\le(1-\tau)N(N-1)/4$, so $M/N\le\sqrt{(1-\tau)/2}$ and $R\le M/T=KM/N\le K\sqrt{(1-\tau)/2}$.
\emph{Part (ii).} $1-r_{\mathrm S}=6\sum\Delta^2/(N(N^2-1))$ and $M(M+1)(2M+1)>2M^3$ (strict as $M>0$
when $R(a)>0$) give $4(M/N)^3<1-r_{\mathrm S}$.

\emph{Tightness.} One family of tables meets both counts with equality. Fix $E\le T$ and list the cells in increasing outcome: the non-leaders of tasks $e>E$ and
all cells but two of each task $e\le E$; then one chosen cell $U_e$ per $e\le E$; then the leaders $W_e$
of tasks $e\le E$; then the other leaders. With $j$ the truth rank, give truth values $\zeta j/N$ below the threshold between
the $U$ and $W$ blocks and $1-\zeta+\zeta j/N$ above. Let the scores follow
the truth except that each $W_e$ is placed just below $U_e$. Tasks $e\le E$ lose $1-O(\zeta)$ each,
the others nothing, and a direct count gives $M(M+1)/2$ discordant pairs and
$\sum\Delta^2=M(M+1)(2M+1)/3$ with $M=E$. With $E=\lfloor\alpha T\rfloor$, $1-\tau\to2(\alpha/K)^2$ and
$1-r_{\mathrm S}\to4(\alpha/K)^3$ as $T\to\infty$, while $R\to\alpha$ as $\zeta\to0$. Letting
$\alpha$ increase to $\min\{1,K\sqrt{(1-\tau)/2}\}$, respectively $\min\{1,K((1-r_{\mathrm S})/4)^{1/3}\}$,
gives both suprema.
\end{proof}

\paragraph{The MMRV.} The bounds of Prop.~\ref{prop:rank} are exact.

\begin{corollary}[the rank-violation statistic, exactly]\label{cor:mmrvexact}
For every $K\ge2$ and $v\ge0$, the supremum of $R(a)$ over outcome tables, score tables with
$\mathrm{MMRV}\le v$ and taskwise maximisers is $\min\{1,Kv\}$. Over score tables that also
have a unique taskwise argmax it is $\min\{1,Kv/2\}$. Both are attained whenever $Kv$ is rational.
\end{corollary}

\begin{proof}[Proof of Corollary~\ref{cor:mmrvexact}]
The upper bounds are Prop.~\ref{prop:rank} and $R\le1$. Take $T$ tasks with column
$(1,0,\dots,0)$. On a fraction $f$ of them, the lost tasks, score $(1,1,0,\dots,0)$, with $a$ picking the tied non-leader; elsewhere scores equal outcomes. On a
lost task the leader's MMRV term is $1$ and every other violated pair has gap zero, so
$\mathrm{MMRV}=f/K$ and $R=f$. For a unique argmax, score $(K-1,K,K-2,\dots,1)$ on the lost tasks;
the terms of the leader and the pick are each $1$, so $\mathrm{MMRV}=2f/K$ and $R=f$. Take
$f=\min\{1,Kv\}$, respectively $\min\{1,Kv/2\}$, with $T$ a denominator of $f$.
\end{proof}

\paragraph{Numerical checks.} In exact arithmetic on $2{,}400$ random outcome tables and correlations, the released code finds no $\varepsilon^\star_p(r)>B_K$; where $B_K<1$ the largest ratio of $\varepsilon^\star_p$ to $B_K$ is $.915$. It checks
Theorem~\ref{thm:rankpooled} on $16{,}983$ random tie-free tables, with no violation.

\paragraph{The correlation needed without a table.} Without an outcome table,
the required correlation is the least $r$ with $V_K(1-r^2)\le\varepsilon$. With $\rho_B$ and $\underline\rho$ the roots of $B_K(\rho)=\varepsilon$ and $\underline B_K(\rho)=\varepsilon$, it lies between $\sqrt{1-\underline\rho}$ and
$\sqrt{1-\rho_B}$ for $K\ge4$, as $V_K$ is nondecreasing in $\rho$; for $K=2$, and for $K=3$ when $\rho_B\le\frac14$, it is $\sqrt{1-\rho_B}$. Every table's exact required correlation is at most
$\sqrt{1-\rho_B}$ by Theorem~\ref{thm:universal}(i).

\begin{proof}[Proof of Theorem~\ref{thm:everyboard}]
\emph{Part (i)} is Corollary~\ref{cor:mmrvexact}. \emph{Part (ii):} the upper bound and
$B_K\le s_K$ are Theorem~\ref{thm:universal}(i) ($B_K<s_K$ when $s_K<1$, and $B_K\le1\le s_K$ otherwise); the lower bound $\min\{1,\underline B_K\}$ for $K\ge4$ and first-order
agreement are Theorem~\ref{thm:universal}(iii); equality for $K=2$, and for $K=3$ with $\rho\le\frac14$, is Theorem~\ref{thm:universal}(ii). \emph{Part (iii)} is
Theorem~\ref{thm:universal}(iv) and (v), with $r_{\mathrm{vac}}(K)=\sqrt{1-\rho_1(K)}$. \emph{Part (iv)}
is the tightness statement of Theorem~\ref{thm:rankpooled}.
\end{proof}

\subsection{The MMRV: proof of Proposition~\ref{prop:rank}, and two comparisons}\label{app:mmrvproof}\label{app:summaries}

\begin{proof}[Proof of Proposition~\ref{prop:rank}]
Let $\mathrm{MMRV}(y,p)\le v$ and let $a$ take a taskwise maximiser of $y$, any one where $y$ ties. Fix a task $t$,
write $i=a(t)$ and $\ell=p_{(1)t}-p_{it}$, and let $A=\{j:p_{jt}>p_{it}\}$. If $\ell=0$ then $A$
is empty and the task contributes nothing to either side, so assume $\ell>0$; then $A$
contains a leader at distance $\ell$ from $p_{it}$.

\emph{Part (i).} Since $y_{it}\ge y_{jt}$ for every $j$, each $j\in A$ has
$[y_{jt}>y_{it}]=0$
while $[p_{jt}>p_{it}]=1$. The pair is misordered in $j$'s summand of \eqref{eq:mmrv}, whether the scores
separate $i$ and $j$ or tie them, and that summand is at least
$p_{jt}-p_{it}$. So the task adds at least
$c^{-}_t(i)=\sum_{j\in A}(p_{jt}-p_{it})\ge\ell$ to $KT\,\mathrm{MMRV}$. (The bound comes from the summands at $j$,
not at $i$: if $y$ ties $i$ with every member of $A$, no pair with $A$ enters $i$'s summand.) Summing over tasks gives
$KTv\ge\sum_t c^{-}_t(a(t))\ge\sum_t\ell_t=T R(a)$.

\emph{Part (ii).} If in addition $y_{it}>y_{jt}$ for every $j\ne i$ and every task, each
$j\in A$ also makes $[y_{it}>y_{jt}]=1\ne0=[p_{it}>p_{jt}]$, so the pair is misordered in $i$'s summand,
which is at least $\max_{j\in A}(p_{jt}-p_{it})=\ell$. The task then adds at least $c^{+}_t(i)=\ell+c^{-}_t(i)\ge2\ell$, and summing gives
$R(a)\le\tfrac{K}{2}v$. The strict maximum cannot be dropped: for
$p=(0,1)^{\top}$ and $y=(1/2,1/2)^{\top}$,
$\mathrm{MMRV}=1/2$, and with the tie broken against the evaluator the regret is $1=Kv>\tfrac{K}{2}v$.

\emph{These lower bounds, the per-task costs, are attained.} In case (ii) the column that puts $a(t)$ strictly above every rival and otherwise
preserves the order of $p$ has cost exactly $c^{+}_t(a(t))$; in case (i) the column that ties $a(t)$ with every policy
above it and preserves the order below has cost exactly $c^{-}_t(a(t))$. So the worst regret compatible with $v$ is a multiple-choice knapsack, the maximum of $R(a)$ over choices of total cost at most $KTv$, solved exactly when $p$ is rational (App.~\ref{app:separable}).
\end{proof}

\paragraph{SIMPLER's tie convention.} SIMPLER's formula counts a tie against the policy the outcome
table ranks lower (on a task with a wrong pick, the pick), not higher as in Eq.~\eqref{eq:mmrv}. Each
task with a wrong pick still contributes at least its loss, and twice its loss when the pick is unique.
So both bounds of Prop.~\ref{prop:rank} and their extremal tables stand, and only the instance bound changes, through the cost of a tied pick (App.~\ref{app:separable}). 

\subsection{An average of within-task correlations}\label{app:within}

Each task's correlation limits how badly that task can be ordered, so routing gets a
guarantee of the same order as the pooled correlation's; deployment and top-$k$ screening get none. Let $c_t$ be the Pearson
correlation of $y_{\cdot t}$ and $p_{\cdot t}$, both assumed nonconstant, and $\hat c=\frac1T\sum_tc_t$.
For a column $v\in[0,1]^K$, $SS(v)=\sum_j(v_j-\bar v)^2$, and
$\Sigma_K=\lfloor K/2\rfloor\lceil K/2\rceil/K$. Let $g_K(c)=\min\{1,\sqrt{2\Sigma_K(1-c_+^2)}\}$ with
$c_+=\max\{c,0\}$. For $K\ge3$, $0<\ell\le1$ and $0\le j\le K-3$, let $SS_j(\ell)$ be the centred sum
of squares of the column $(1,1-\ell,(1-\ell/2)^{\times j},0^{\times(K-2-j)})$, and for $c\in(0,1)$ let
$\underline g_K(c)$ be the largest $\ell\in(0,1]$ with $\ell^2/2\le(1-c^2)\max_{j\le K-3}SS_j(\ell)$.

\begin{proposition}[the average of within-task correlations]\label{prop:within}
Fix $K\ge2$.
\emph{(i)} If $\hat c\ge\bar c$, every taskwise maximiser $a$ has $R(a)\le g_K(\bar c)$.
\emph{(ii)} For $K\ge3$, $\bar c\in(0,1)$ and every $T$, some outcome and score tables have $c_t\ge\bar c$ on
every task and, with ties broken against the evaluator, regret $\underline g_K(\bar c)$; and
$\underline g_K(\bar c)/g_K(\bar c)\to1$ as $\bar c\to1$.
\emph{(iii)} For every $K\ge2$, $T\ge2$, $1\le k\le K-1$ and $\delta\in(0,1)$, some pair of tables has
$c_t=1$ on every task and an MMRV of $0$, so per-task regret zero, while the evaluator's top $k$
policies by mean score all trail the best policy mean by $(T-1-\delta)/T$.
\end{proposition}

\begin{proof}
\emph{Part (i).} Fix a task, its column $v=p_{\cdot t}$, its pick $i=a(t)$ and loss
$\ell=\max_jv_j-v_i$. The score column ranks $i$ first, so Prop.~\ref{prop:cone} applied to the
one-column table gives $c_t\le\sqrt{1-d_t(i)/SS(v)}$ when $c_t>0$. By Lemma~\ref{lem:rung}, $d_t(i)\ge\ell^2/2$, and
$SS$ is convex, so it peaks at a vertex of the cube, where with $n$ entries equal to one it is
$n(K-n)/K\le\Sigma_K$. Hence $\ell^2\le2\,d_t(i)\le2\,SS\,(1-c_t^2)\le2\Sigma_K(1-c_t^2)$, so
$\ell\le g_K(c_t)$; for $c_t\le0$ this holds as $g_K(c_t)=\min\{1,\sqrt{2\Sigma_K}\}=1$, since
$2\Sigma_K\ge1$. The function $g_K$ is nonincreasing and concave on $[-1,1]$ (its arc is flat at $c=0$), so by Jensen's inequality
$R(a)=\frac1T\sum_t\ell_t\le\frac1T\sum_tg_K(c_t)\le g_K(\hat c)\le g_K(\bar c)$.

\emph{Part (ii).} Let $\ell=\underline g_K(\bar c)$ and $j\le K-3$ attain the maximum, and take the
column $v=(1,1-\ell,(1-\ell/2)^{\times j},0^{\times(K-2-j)})$ with the second entry chosen. Setting
the leader and the chosen entry to $1-\ell/2$ gives a column that ranks the chosen entry first, at
squared distance $\ell^2/2$, the floor of Lemma~\ref{lem:rung}; so it is the projection, and it is
nonconstant because it keeps a zero. As $\ell^2/2\le(1-\bar c^2)SS_j(\ell)<SS_j(\ell)$,
Prop.~\ref{prop:cone} gives it the correlation $\sqrt{1-\ell^2/(2SS_j(\ell))}\ge\bar c$ with $v$. Repeat
this pair of columns on every task. The projection ties the chosen entry with the leader, and the
tie broken against the evaluator loses $\ell$ on every task. As $\bar c\to1$, $\underline g_K\le g_K\to0$, so the column tends to $j+2$ ones and $K-2-j$ zeros, and taking $j+2=\lceil K/2\rceil$, admissible for $K\ge3$, gives $\max_jSS_j\to\Sigma_K$. Once $\underline g_K<1$, the defining inequality holds with equality,
$\underline g_K^2=2(1-\bar c^2)\max_jSS_j(\underline g_K)$, so $\underline g_K/g_K\to1$.

\emph{Part (iii).} Take $k$ copies of a policy A with outcome $1$ on task $1$ and $0$ elsewhere, a
policy B with $1-\delta$ on task $1$ and $1$ elsewhere, and the remaining $K-k-1$ policies at $0$
everywhere; every column is nonconstant. Score each task by its outcomes, multiplied by
$\Lambda>(T-1)/\delta$ on task $1$. Each $c_t=1$ and every pick is right, but each copy of A has mean
score $\Lambda/T$, above B's $(\Lambda(1-\delta)+T-1)/T$, so the top $k$ are the copies, whose mean
outcome $1/T$ trails B's $(T-\delta)/T$ by $(T-1-\delta)/T$. These scores order every task exactly as
the outcomes do, so their MMRV is $0$.
\end{proof}

For even $K$, $2\Sigma_K=K/2$, so the leading term $\sqrt{2\Sigma_K(1-\bar c^2)}$ equals, at $r=\bar c$,
the pooled correlation's $\sqrt{K(1-r^2)/2}$. Table~\ref{tab:census-within} applies the proposition to
the averages published in the robot papers of App.~\ref{app:census}. For $K=2$ every $c_t$ is $\pm1$, so at most a share $(1-\bar c)/2$ of the tasks is misordered and $R(a)\le(1-\bar c)/2$, tighter than $g_2$; \texttt{evalceil/card.py} uses this.

\subsection{Proof of Theorem~\ref{thm:noise}: guarantees on the scale of the errors}\label{app:noise}

The idea of the proof is that the correlation sees an error through its square, while a decision
changes only when an error crosses the gap between two policies. Part (i) is an exact identity for
$1-r^2$; parts (ii) and (iii) bound how often, and by how much, an error of size $\sigma$ crosses a gap.
Write $x$, $z$ and $w$ for the tables $p$, $Z$ and $y$ as vectors in $\mathbb R^{KT}$, $\mathbf 1$ for
the all-ones vector, $u=x-\bar x\mathbf 1$, so that $\lVert u\rVert^2=S$ and $u\perp\mathbf 1$, and
$\tilde w=w-\bar w\mathbf 1=u+\sigma(z-\bar z\mathbf 1)$.

\begin{proof}[Proof of Theorem~\ref{thm:noise}]
\emph{Part (i).} Let $P_{\mathbf 1}$, $P_u$ be the projections onto $\mathbf 1$ and $u$, $\Pi=I-P_{\mathbf 1}-P_u$ the projection onto the complement of their span, and $W=\lVert\Pi z\rVert^2/S$. As $r=\langle u,\tilde w\rangle/(\lVert u\rVert\lVert\tilde w\rVert)$, $1-r^2=\lVert\tilde w-P_u\tilde w\rVert^2/\lVert\tilde w\rVert^2$. Since $P_uu=u$ and $P_u\mathbf 1=0$, $\tilde w-P_u\tilde w=\sigma(z-\bar z\mathbf 1-P_uz)=\sigma\Pi z$, so $1-r^2=\sigma^2\lVert\Pi z\rVert^2/\lVert y-\bar y\rVert_F^2$, the residual share of the least-squares regression of $y$ on $p$. For fixed $z$, $\lVert\tilde w\rVert^2\to S$ as $\sigma\to0$, so $\rho/\sigma^2\to W$. As $SW$ is chi-squared with $KT-2\ge1$ degrees of freedom, $W>0$ almost surely and $\E W=(KT-2)/S$. The guarantee from
$r$ and $K$ is $V_K(\rho)$ of App.~\ref{app:universal}, and $V_K(\rho)=\sqrt{K\rho/2}+O(\rho)$ for every $K$ by Theorem~\ref{thm:universal}(iii). Hence
$V_K(\rho)/\sigma\to\sqrt{KW/2}$.

\emph{Part (ii).} Scores tie with probability zero, so almost surely every task has a unique top
policy. Let $Q(x)=\Pr[N(0,1)>x]$. A pair $i,j$ on task $t$ with $p_{it}>p_{jt}$ adds to Eq.~\eqref{eq:mmrv} only if it is \emph{misordered}, $y_{it}\le y_{jt}$, that is $\sigma(Z_{jt}-Z_{it})\ge p_{it}-p_{jt}\ge\gamma$. As $Z_{jt}-Z_{it}\sim N(0,2)$, this has probability
$Q\bigl((p_{it}-p_{jt})/(\sigma\sqrt2)\bigr)\le\frac12e^{-\gamma^2/(4\sigma^2)}$; pairs with equal
outcomes contribute $0$. A union bound over the $N$ pairs gives the bound on $\Pr[\mathrm{MMRV}>0]$.
If no pair is misordered, a policy above the pick in $p$ would also be above it in $y$, so the pick is
a best policy on every task and the regret is $0$. The regret is at most $1$, and the MMRV's guarantee
$\frac K2\mathrm{MMRV}$ at most $\frac K2$, since every gap is at most $1$; and $e^{-a/\sigma^2}\le\sigma^2/(ea)$
for $a>0$, as $se^{-s}\le e^{-1}$. With $a=\gamma^2/4$ this gives the two expectations.

\emph{Part (iii).}  The term of policy $i$ on a task is at most the sum over $j\ne i$ of
$\lvert g\rvert\,\mathbf 1[\text{$i,j$ misordered}]$, with $g=p_{it}-p_{jt}$. Given $g$, the pair is
misordered with probability $Q(\lvert g\rvert/(\sigma\sqrt2))$, and $g$, a difference of
independent variables one of which has density at most $f$, has density at most $f$. So each pair's contribution has expectation at most $f\int_{\mathbb R}\lvert g\rvert\,Q(\lvert g\rvert/(\sigma\sqrt2))\,dg
=4f\sigma^2\int_0^\infty s\,Q(s)\,ds=f\sigma^2$, since $\int_0^\infty sQ(s)\,ds=\frac14$.
Summing over the $K-1$ partners $j$ and averaging over the $KT$ policy--task terms gives $\E[\mathrm{MMRV}]\le(K-1)f\sigma^2$. With a
unique top policy on every task, Prop.~\ref{prop:rank}(ii) gives $R(a)\le\frac K2\mathrm{MMRV}$, and the
expectations follow.
\end{proof}

\paragraph{Figure~\ref{fig:noise}.} \texttt{sims/noise\_scale.py} draws $4{,}000$ score tables
$y=p+\sigma Z$ per outcome table and $\sigma$, computes the correlation's value-only bound (the lower
end of Theorem~\ref{thm:everyboard}(ii), and, by monotonicity, $B_3(\frac14)=.5$ for $K=3$ above $\rho=\frac14$) and the
MMRV's, $Kv/2$, and integrates the regret's expectation over the noise by quadrature. It asserts the identity of part (i) on every draw and the bound of part (ii)
against the observed frequency.

%% file: table_needed.tex
\begin{table}[b]
\centering\footnotesize\setlength{\belowcaptionskip}{3pt}
\caption{Correlation needed to guarantee a per-task regret of at most a given tolerance, by number of policies $K$ (Theorem~\ref{thm:everyboard}). At or above this value, no outcome table allows a larger regret; $.0001$ below it, some table does. Values are rounded up.}
\label{tab:needed}
\begin{tabular}{@{}cccc@{}}
\toprule
$K$ & 2 points & 5 points & 10 points\\
\midrule
2 & $.99980$ & $.99862$ & $.99389$ \\
3 & $.99987$ & $.99911$ & $.99621$ \\
4 & $.99990$ & $.99935$ & $.99725$ \\
5 & $.99992$ & $.99949$ & $.99785$ \\
6 & $.99994$ & $.99957$ & $.99823$ \\
8 & $.99995$ & $.99968$ & $.99870$ \\
10 & $.99996$ & $.99975$ & $.99897$ \\
\bottomrule
\end{tabular}
\end{table}

%% file: app_corpus.tex
\section{The corpus and the reported statistics}\label{app:corpus}

\subsection{The corpus}

The corpus holds eleven validations: eleven evaluators' score tables, each with the real outcome table it
is validated against. They come from nine distinct outcome tables and eight \emph{source tables},
released data files transcribed from a paper's printed table. A source table
can hold several outcome tables (Ctrl-World's holds both of its criteria), and an outcome table can
serve several evaluators (SIMPLER's Google Robot table and AgentRewardBench's). Rates are used as
printed, divided by the trial count where the source prints successes (WorldGym, AutoEval) and by one
hundred where it prints a hundred-point score (AgentRewardBench, WMT23). The only rounding is of chrF,
divided by one hundred and rounded to four places. Every rate is thus an exact fraction, and
\texttt{code/regenerate.py} computes every corpus quantity in exact rational arithmetic.

\paragraph{The eight source tables.} \emph{WorldGym} \citep{worldgym}: three policies in a learned
video world model on $17$ tasks, against real Bridge outcomes from OpenVLA's evaluation, with half
credit, so every entry is a multiple of $1/20$. \emph{SIMPLER} \citep{simpler}: real policies re-run in
simulation; Google Robot with two evaluator columns (visual matching, variant aggregation), and
WidowX. \emph{Ctrl-World} \citep{ctrlworld}: three policies in a controllable world model, scored on
instruction following and on task success, each its own outcome table; entries are multiples of
$1/20$. \emph{AutoEval} \citep{autoeval}: six policies scored autonomously (evaluator) and under human
supervision (outcome) on the same hardware; a second file pairs SIMPLER, as evaluator, with the
supervised outcomes of the four tasks it simulates. \emph{AgentRewardBench} \citep{agentrewardbench}:
four web agents on four of its five benchmarks (Llama 3.3 lacks VisualWebArena) against experts, with a GPT-4o judge and a rule-based checker. \emph{WMT23}
\citep{kocmi-etal-2023-findings,wmt23newssystems}: the nine systems present in all eight language pairs
of the general task, without the human reference; the outcome is the human DA+SQM score and the
evaluator chrF \citep{popovic2015chrf}.

\paragraph{Files, trials and ties.} The source files are in \texttt{code/data/}; $K$ and $T$ are in
Table~\ref{tab:eleven}. Real trials per cell: $10$ on
WorldGym, $75$, $60$, $54$ and $27$ by task on SIMPLER's Google Robot, $24$ on WidowX and $50$ on
AutoEval; the others state none. Tasks with a tie for the best real outcome: two on Ctrl-World's
instruction table, one on its success table and one on AgentRewardBench's. Tasks with a tie for the evaluator's top score: two for Ctrl-World's success evaluator and one for AgentRewardBench's rule-based
checker; none elsewhere.

\paragraph{What each source publishes.} Only WorldGym prints a pooled correlation, $.78$, which we recompute as $.785$. SIMPLER and
AutoEval average a within-task coefficient over tasks (on AutoEval's autonomous table, $.942$ against a
pooled $.992$); AutoEval's printed MMRV, $.015$, matches our $.0147$. Ctrl-World prints no correlation,
AgentRewardBench precision and recall, and WMT23 compares metrics in a separate shared task.

\paragraph{Reservations.} WorldGym prints a world-model mean of $15.5\%$ for RT-1-X, while its own per-task counts give $15.59\%$; we use the counts. Two of SIMPLER WidowX's twelve real rates, $.233$ and $.433$, cannot come from $24$ trials under any rounding; we use them as printed. In SIMPLER's
Google Robot file one cell, RT-1 (Begin) on the drawer task, completed $2$ of its $54$ planned trials, and one completed count is blank; we use the
planned counts. That cell's real rate is $0$, which resamples to $0$ at any count, and at the completed counts the $95\%$ box of App.~\ref{app:hinge} has the same hinge (\texttt{sims/anchor\_depth.py} asserts it). Four of the eight source tables, AutoEval's two, AgentRewardBench's and WorldGym's real column, are compared cell by cell with values copied from the source; the others are checked only for shape, completeness and, where a trial count is stated, agreement with it (\texttt{sims/corpus\_audit.py}). AutoEval's two outcome tables share their real cells, so a statement about both
is one observation. Sampling simulations treat the printed rates as true and trials as independent. No sampling statement is made about Ctrl-World, AgentRewardBench or WMT23.

\subsection{The reported statistics}\label{app:census}

We collected every validation we could find in papers that validate a robot-policy evaluator against
real outcomes and print a Pearson correlation or an MMRV with the number of policies ranked. Eleven
papers from 2024 to 2026 qualify, with $52$ validations, $14$ of them from SimFoundry
(Table~\ref{tab:census}); \texttt{data/census\_published.csv} names the table or figure each value is
read from. A validation is one evaluator against one real table, either on a single task (SIMPLER,
WorldEval, SimFoundry, PolaRiS) or pooled over a setting (REALM, WEAVER). We leave out tables that do
not state $K$, correlations across tasks for one policy, and rank correlations. Averages of per-task
correlations are not entries; Table~\ref{tab:census-within} converts every such group these papers print.
PolaRiS \citep{polaris2025} prints only such averages, so it enters through the SimFoundry paper,
which reports it as an evaluator. This collection omits the corpus's outcome tables, analysed exactly above, except SIMPLER's Google Robot per-task correlations. Evaluator names follow the sources.

\input{table_census}

\paragraph{What it shows.} No correlation guarantees five points. Of the $51$ printed correlations,
$28$ are computed on a single task, and pricing them over one-task tables changes no conclusion. Even
the highest, $.995$, a single-task value, leaves open a per-task regret of at least $.144$. For $13$ of
the $51$, $2$ of them negative, the guarantee is $1$: the correlation is vacuous ($12$ of the $13$ are single-task). The MMRV guarantees five points on $7$ validations ($14$ if the evaluator's top policy is unique
on every task) and zero regret on $6$. Of the $47$ that print both statistics, the MMRV's value-only
bound is strictly tighter than any the correlation can support on $32$ ($44$ with a unique top),
looser than the correlation's upper bound $B_K$ on $4$, both vacuous on $9$ and not ordered on $2$.

\begin{table}[t]
\centering\tabfont\setlength{\belowcaptionskip}{3pt}\setlength{\tabcolsep}{4pt}
\caption{Within-task correlations printed in the robot papers of App.~\ref{app:census}, their mean (AutoEval prints only this), and the per-task regret it leaves open on outcome tables with that $K$ (Prop.~\ref{prop:within}). \emph{At least}: $\underline g_K$ at the average, attained on some outcome table, rounded down; \emph{at most}: $g_K$, rounded up. An entry of $1$ is vacuous.}
\label{tab:census-within}
\begin{tabular}{@{}lrlrcc@{}}
\toprule
evaluator & $K$ & per-task correlations & average & at least & at most\\
\midrule
SIMPLER, visual matching & $6$ & $.976, .855, .942$ & $.924$ & $.526$ & $.661$\\
SIMPLER, variant aggregation & $6$ & $.960, .887, .486$ & $.778$ & $.823$ & $1$\\
SIMPLER, validation action MSE & $6$ & $.464, .230, .231$ & $.308$ & $1$ & $1$\\
WorldEval, world model & $4$ & $.958, .887, .980$ & $.942$ & $.403$ & $.476$\\
WorldEval, real-to-sim baseline & $4$ & $.328, .442, .465$ & $.412$ & $1$ & $1$\\
SimFoundry, three policies & $3$ & $.883, .915, .910$ & $.903$ & $.431$ & $.497$\\
SimFoundry, five policies & $5$ & $.960, .907, .995, .810$ & $.918$ & $.501$ & $.615$\\
PolaRiS, five policies & $5$ & $.480, -.396, .512, -.037$ & $.140$ & $1$ & $1$\\
AutoEval, autonomous scorer & $6$ & not printed & $.942$ & $.471$ & $.582$\\
\bottomrule
\end{tabular}
\end{table}

AutoEval's printed average, $.942$, leaves open $.471$ to $.582$; the pooled correlation we recompute, $.992$, has an instance bound of $.096$ (Table~\ref{tab:eleven}).

%% file: table_census.tex
\begin{table}[t]
\centering\renewcommand{\arraystretch}{0.86}\tabfont\setlength{\belowcaptionskip}{3pt}\setlength{\tabcolsep}{1.6pt}
\caption{Per-task regret guarantees for every published validation in eleven papers from 2024 to 2026
that validate a robot-policy evaluator and print a Pearson correlation $r$ or an MMRV $v$ beside $K$
(Theorem~\ref{thm:everyboard}). \emph{By $r$}: the worst per-task regret the correlation leaves open, in one of four forms.
For $K\ge4$ it is an interval from $\min\{1,\underline B_K\}$ to $B_K$, rounded outward.
For $K=2$, and for $K=3$ with $1-r^2\le\frac14$, it is the exact value $B_K$, rounded up.
For $K=3$ with $1-r^2>\frac14$ it is an interval from $B_3(\frac14)=.5$, since a lower $r$ admits more score tables, to $B_3$.
It is $1$ at or below $r_{\mathrm{vac}}(K)$ and for a negative $r$. SIMPLER, WorldEval, SimFoundry and PolaRiS print each correlation on a single task. There the worst case is over one-task tables, and its lower end is $\underline g_K$ of Prop.~\ref{prop:within}; this changes only the rows marked $\dagger$. \emph{By MMRV}: $\min\{1,Kv\}$, which always holds;
\emph{unique}: $\min\{1,Kv/2\}$, which holds when the evaluator's top policy is unique on every task; both rounded up.
An entry of $1$ is vacuous. n.p.: not printed.}
\label{tab:census}
\begin{tabular}{@{}llcccccc@{}}
\toprule
 & & & \multicolumn{2}{c}{printed} & \multicolumn{3}{c}{worst per-task regret left open} \\
\cmidrule(lr){4-5}\cmidrule(lr){6-8}
source & evaluator, outcome table & $K$ & $r$ & MMRV & by $r$ & by MMRV & unique \\
\midrule
SIMPLER \citep{simpler} & visual matching, pick coke can & 6 & $.976$ & $.031$ & $.323$ to $.348$ & $.186$ & $.093$ \\
 & visual matching, move near & 6 & $.855$ & $.111$ & $.688$ to $.862$ & $.666$ & $.333$ \\
 & visual matching, drawer & 6 & $.942$ & $.027$ & $.471$ to $.531$ & $.162$ & $.081$ \\
 & variant aggregation, pick coke can & 6 & $.960$ & $.084$ & $.403$ to $.444$ & $.504$ & $.252$ \\
 & variant aggregation, move near & 6 & $.887$ & $.111$ & $.621$ to $.748$ & $.666$ & $.333$ \\
 & variant aggregation, drawer & 6 & $.486$ & $.235$ & $1$ & $1$ & $.705$ \\
 & validation action MSE, pick coke can & 6 & $.464$ & $.412$ & $1$ & $1$ & $1$ \\
 & validation action MSE, move near & 6 & $.230$ & $.408$ & $1$ & $1$ & $1$ \\
 & validation action MSE, drawer & 6 & $.231$ & $.306$ & $1$ & $1$ & $.918$ \\
\addlinespace[2pt]
WorldEval \citep{worldeval} & WorldEval, place cup & 4 & $.958$ & $.000$ & $.349$ to $.357$ & $.000$ & $.000$ \\
 & WorldEval, strike block & 4 & $.887$ & $.133$ & $.538$ to $.568$ & $.532$ & $.266$ \\
 & WorldEval, handover block & 4 & $.980$ & $.000$ & $.251$ to $.254$ & $.000$ & $.000$ \\
 & real-to-sim, place cup & 4 & $.328$ & $.332$ & $1$ & $1$ & $.664$ \\
 & real-to-sim, strike block & 4 & $.442$ & $.266$ & $1$ & $1$ & $.532$ \\
 & real-to-sim, handover block & 4 & $.465$ & $.187$ & $1$ & $.748$ & $.374$ \\
\addlinespace[2pt]
Veo & Veo, nominal scenarios & 8 & $.88$ & $.03$ & $.701$ to $.936$ & $.240$ & $.120$ \\
\citep{veo2025} & Veo, background & 5 & $.91$ & $.00$ & $.529$ to $.589$ & $.000$ & $.000$ \\
 & Veo, small distractor & 5 & $.86$ & $.10$ & $.638$ to $.743$ & $.500$ & $.250$ \\
 & Veo, large distractor & 5 & $.77$ & $.14$ & $.791$ to $1$ & $.700$ & $.350$ \\
 & Veo, object & 5 & $.56$ & $.15$ & $1$ & $.750$ & $.375$ \\
\addlinespace[2pt]
REALM \citep{realm2025} & pooled & 3 & $.92$ & $.118$ & $.397$ & $.354$ & $.177$ \\
 & default & 3 & $.88$ & $.166$ & $.476$ & $.498$ & $.249$ \\
 & object pose & 3 & $.93$ & $.115$ & $.374$ & $.345$ & $.173$ \\
 & camera pose & 3 & $.89$ & $.253$ & $.457$ & $.759$ & $.380$ \\
\addlinespace[2pt]
Video WM \citep{tseng2025videowm} & Cosmos, Bridge & 3 & $.687$ & $.171$ & $.500$ to $.793$ & $.513$ & $.257$ \\
 & IRASim, Bridge & 3 & $.613$ & $.611$ & $.500$ to $.937$ & $1$ & $.917$ \\
\addlinespace[2pt]
dWorldEval \citep{dworldeval} & dWorldEval, RoboTwin & 3 & $.927$ & $.033$ & $.381$ & $.099$ & $.050$ \\
 & dWorldEval without history, RoboTwin & 3 & $.863$ & $.033$ & $.500$ to $.506$ & $.099$ & $.050$ \\
\addlinespace[2pt]
H2RBench \citep{h2rbench} & simulation, robot-only demonstrations & 4 & $.971$ & $.014$ & $.296$ to $.301$ & $.056$ & $.028$ \\
 & simulation, robot and human demonstrations & 4 & $.894$ & $.060$ & $.523$ to $.550$ & $.240$ & $.120$ \\
\addlinespace[2pt]
SO-101 Bench \citep{so101bench} & digital twin, policy level, unseen real environment & 5 & $.90$ & $.000$ & $.553$ to $.621$ & $.000$ & $.000$ \\
\addlinespace[2pt]
RoboWorld \citep{roboworld} & GPT-4o judge, RoboWorld Score & 8 & $.989$ & n.p. & $.260$ to $.281$ & n.p. & n.p. \\
 & Gemini judge, RoboWorld Score & 8 & $.944$ & n.p. & $.517$ to $.620$ & n.p. & n.p. \\
 & GPT-4o judge, success rate & 8 & $.901$ & n.p. & $.650$ to $.838$ & n.p. & n.p. \\
 & Gemini judge, success rate & 8 & $.876$ & n.p. & $.710$ to $.954$ & n.p. & n.p. \\
\addlinespace[2pt]
WEAVER \citep{weaver} & Ctrl-World, DROID & 2 & $.552$ & $.215$ & $.696$ & $.430$ & $.215$ \\
 & WEAVER, DROID & 2 & $.563$ & $.155$ & $.684$ & $.310$ & $.155$ \\
 & WEAVER finetuned, DROID & 2 & $.863$ & $.035$ & $.370$ & $.070$ & $.035$ \\
\addlinespace[2pt]
SimFoundry & SimFoundry, Stack Dishware\smash{$^\dagger$} & 3 & $.883$ & $.000$ & $.469$ to $.470$ & $.000$ & $.000$ \\
\citep{simfoundry} & SimFoundry, Store Marker\smash{$^\dagger$} & 3 & $.915$ & $.000$ & $.405$ to $.407$ & $.000$ & $.000$ \\
 & SimFoundry, Throw Away Trash\smash{$^\dagger$} & 3 & $.910$ & $.067$ & $.416$ to $.418$ & $.201$ & $.101$ \\
 & SimFoundry, Serve Fruits\smash{$^\dagger$} & 5 & $.960$ & $.016$ & $.368$ to $.399$ & $.080$ & $.040$ \\
 & SimFoundry, Cup in Bowl\smash{$^\dagger$} & 5 & $.907$ & $.016$ & $.529$ to $.599$ & $.080$ & $.040$ \\
 & SimFoundry, Marker in Cup\smash{$^\dagger$} & 5 & $.995$ & $.008$ & $.144$ to $.150$ & $.040$ & $.020$ \\
 & SimFoundry, Clear Table\smash{$^\dagger$} & 5 & $.810$ & $.016$ & $.718$ to $.891$ & $.080$ & $.040$ \\
 & PolaRiS, Stack Dishware & 3 & $.500$ & $.200$ & $1$ & $.600$ & $.300$ \\
 & PolaRiS, Store Marker\smash{$^\dagger$} & 3 & $.822$ & $.053$ & $.571$ to $.573$ & $.159$ & $.080$ \\
 & PolaRiS, Throw Away Trash & 3 & n.p. & $.253$ & n.p. & $.759$ & $.380$ \\
 & PolaRiS, Serve Fruits & 5 & $.480$ & $.288$ & $1$ & $1$ & $.720$ \\
 & PolaRiS, Cup in Bowl & 5 & $-.396$ & $.280$ & $1$ & $1$ & $.700$ \\
 & PolaRiS, Marker in Cup & 5 & $.512$ & $.176$ & $1$ & $.880$ & $.440$ \\
 & PolaRiS, Clear Table & 5 & $-.037$ & $.352$ & $1$ & $1$ & $.880$ \\
\bottomrule
\end{tabular}
\end{table}

%% file: app_audits.tex
\section{The four audits}\label{app:audits}

For each audit, a criteria
file in \texttt{code/} fixes the population, the table construction, the tolerances, the statistics and
the hypotheses with their deciding thresholds. It was written before any evaluator score was read or
any statistic computed; no criteria file is publicly time-stamped. Table~\ref{tab:wmthyp} lists every hypothesis
and its outcome. Every audit computes, for each validation, the pooled correlation $r$, the MMRV and
pairwise accuracy within tasks, and the instance bound of each for routing:
Eq.~\eqref{eq:exchange} for $r$, Prop.~\ref{prop:separable} for the other two.

\begin{table}[t]
\centering\tabfont\setlength{\belowcaptionskip}{3pt}
\caption{Every hypothesis registered for the four audits, each in its \texttt{code/CRITERIA\_*.md} file before any evaluator score was read: WMT (W), DataDecide (D), PPE (R) and NAS-Bench-201 (N). \emph{Threshold}: the value that decides the hypothesis, fixed in advance. Guarantees are for routing unless stated; five points is $\varepsilon=.05$. $^\ast$: a part that follows closely from Theorem~\ref{thm:everyboard} at the audit's $K$, a low-risk check (this grouping is ours, made after the audits; the criteria files do not mark it); all $5$ held. Of the $20$ other parts, $14$ held and $6$ failed.}
\label{tab:wmthyp}
\begin{tabularx}{\linewidth}{@{}l>{\raggedright\arraybackslash}Xll@{}}
\toprule
& hypothesis and threshold & result & verdict\\
\midrule
W-H1a$^\ast$ & from $r$ and $K$ alone, no validation certifies five points & $0$ & held\\
W-H1b$^\ast$ & at least half the correlations are at or below $r_{\mathrm{vac}}(K)$ & $673$ of $702$ & held\\
W-H2a & MMRV's instance bound below $r$'s on at least three quarters & $628$ of $702$ & held\\
W-H2b & pairwise accuracy's below $r$'s on at least three quarters & $59$ of $702$ & failed\\
W-H3a & $r$'s instance bound at most five points on at most a tenth & $21$ of $702$ & held\\
W-H3b & $\Phi(p)$ (Sec.~\ref{sec:exchange}) above five points on at least half the tables & $10$ of $33$ & failed\\
W-H4a & median $\tau_b(r,-\text{regret})$ over tables at most $.3$ & $.32$ & failed\\
W-H4b & median $\tau_b$ for the MMRV above that for $r$ & $.52$ against $.32$ & held\\
W-H5 & median guarantee over regret larger for $r$ than for the MMRV & $6.0$ against $3.4$ & held\\
W-H6a & year-$t$ guarantee of $r$ bounds year-$(t+1)$ regret in at least $95\%$ & $53$ of $55$ & held\\
W-H6b & the same for the MMRV & $39$ of $55$ & failed\\
W-H6c & year-$t$ regret bounds year-$(t+1)$ regret in at most $75\%$ & $16$ of $55$ & held\\
\midrule
D-H1$^\ast$ & at least half the correlations at or below $r_{\mathrm{vac}}(25)$ & $10$ of $13$ & held\\
D-H2 & MMRV's instance bound below $r$'s on at least three quarters & $13$ of $13$ & held\\
D-H3 & $\tau_b$ for the MMRV above that for $r$ & $.87$ against $.79$ & held\\
D-H4 & no size of 150M or fewer certifies deployment within one point by decision accuracy & $0$ of $10$ & held\\
\midrule
R-H1$^\ast$ & every correlation at or below $r_{\mathrm{vac}}(32)$ & $70$ of $70$ & held\\
R-H2 & MMRV's instance bound below $r$'s on at least three quarters & $1$ of $70$ & failed\\
R-H3 & median $\tau_b$ for the MMRV above that for $r$ & $.66$ against $.64$ & held\\
R-H4 & median $\tau_b$ for pairwise accuracy above that for $r$ & $.86$ against $.64$ & held\\
R-H5 & no statistic certifies five points on any validation & $0$ & held\\
\midrule
N-H1$^\ast$ & from $r$ and $K$ alone no proxy guarantees any regret below one & $0$ of $13$ & held\\
N-H2 & with the table no correlation certifies one point, per dataset or for deployment & $0$ of $13$ & held\\
N-H3 & the best proxy's realised regret is below five points & $.020$ & held\\
N-H4 & $\tau_b$ of Precision@100 with screen regret not above Spearman's & $.79$ against $.22$ & failed\\
\bottomrule
\end{tabularx}
\end{table}

\subsection{Translation metrics: WMT22 to WMT25}\label{app:wmtaudit}

The metrics shared task of the Conference on Machine Translation publishes many
evaluators scored on the same human-judged tables, with ten to twenty systems each
\citep{freitag2022wmt22metrics,freitag2023wmt23metrics,freitag2024wmt24metrics,lavie2025wmt25metrics}.

\paragraph{Data and construction.} The data are the mt-metrics-eval archive \citep{mtmetricseval} and
its metadata for each table: standard reference, standard human score and outlier systems. The
population is every language pair of WMT22 to WMT25 whose standard human score is MQM, error annotation
by professional translators \citep{freitag2021experts}, or ESA, error span annotation on a $0$--$100$
scale \citep{kocmi2024esa}: eleven MQM and twenty-two ESA tables. The policies are the systems,
including human translations other than the standard reference and excluding outliers. The tasks are
the domains (without the canary domain), each restricted to the segments every system has a human
score on. An MQM error score $m$ gives the outcome $\max\{0,1-m/25\}$ ($25$ marks a non-translation)
and an ESA score $s$ gives $s/100$; a cell is the mean outcome over the domain's segments, rounded to
four places. The evaluators are the year's primary and baseline metrics, read with the standard
reference or, for a reference-free metric, the source. A metric enters a table only if it scores every
kept segment of every system; its cell is the mean of its segment scores. WMT25 English-Maasai has no
such metric, so the audit has $32$ tables, with $10$ to $20$ systems, two to five domains and $13$ to
$32$ metrics each: $702$ validations. Pairwise accuracy is computed within domains with the sign rule of \citet{kocmi2021ship}, leaving out system pairs the outcome table ties; its per-option cost is the
number of systems above the option (Prop.~\ref{prop:pairwise}). \texttt{tests/test\_wmt\_audit.py}
checks each solver against enumeration, with code independent of the library.

\paragraph{Outcomes on any bounded scale.} If $p'=\alpha p+\beta\mathbf 1$ with $\alpha>0$, correlations,
pairwise accuracy and orders are unchanged and regrets, gaps and the MMRV scale by $\alpha$; so do all
bounds, with an MMRV in the outcome's own units entering as $v/\alpha$.

\paragraph{The registered hypotheses.} Eight of the twelve parts held (Table~\ref{tab:wmthyp}); the
four failures are stated here, as the criteria file requires. \emph{W-H2b:} pairwise accuracy's instance bound is below the correlation's on $59$ of $702$ validations, not three quarters, because pairwise accuracy counts the systems above a wrong pick, not their gaps. \emph{W-H3b:} $\Phi(p)$
exceeds five points on $10$ of the $33$ tables (English-Maasai included: $\Phi$ needs no metric), not half; on the other $23$, Theorem~\ref{thm:levels} does not exclude certifying deployment. \emph{W-H4a:} the correlation's median $\tau_b$ is $.32$, above the threshold of $.3$.
\emph{W-H6b:} in the $55$ cases where a metric appears under the same name in consecutive years on the
same language pair, its year-$t$ MMRV guarantee bounds its year-$(t+1)$ regret in $39$, below $95\%$,
while the looser correlation's does in $53$. The years differ in systems and domains, so no theorem
promises such a transfer.

\paragraph{From the statistic alone.} With $10$ to $20$ systems, $r_{\mathrm{vac}}(K)$ runs from $.853$ to $.927$ (rounded down), and $673$ of the $702$ correlations lie at or below it, $57$ of them at
or below zero; none guarantees five points. 

\paragraph{With the tables.} The MMRV's instance bound is strictly below the correlation's on $628$
validations and equal on the other $74$. The correlation certifies routing on $21$, all on
WMT22 English-German, where no choice loses five points; the MMRV certifies it on $122$. The
correlation's guarantee never falls below the single best policy's regret, and the metrics' own
per-domain choice does worse than the single best policy on $587$ validations and better on only $86$.
Over the $698$ with a positive realised regret, the median ratio of guarantee to realised regret is
$6.0$ for the correlation and $3.4$ for the MMRV.

\paragraph{Deployment and top-$k$ screening.} The pooled correlation certifies deployment on $31$ validations and a
top-2 screen on $117$. The system-level statistics that WMT22--23 report, the Pearson correlation and pairwise accuracy of the $K$ system means, computed here on all $702$ validations, do better as instance bounds for deployment: they certify it on $115$ and $77$ validations, and order the metrics by deployment regret with
median $\tau_b$ of $.40$ and $.57$, against $.28$ for the pooled correlation.

\paragraph{The average of within-task correlations (exploratory).} On every audit table we also computed each evaluator's mean over tasks of its within-task Pearson correlation and its
$\tau_b$ with minus the realised regret (\texttt{sims/within\_tau.py}). Constant columns are left out, and with them the sentinel metrics that read only the source or the reference. It is $.49$ on WMT and $.82$ on PPE (medians over tables), $.87$ on DataDecide and $.30$ on NAS, against $.32$, $.64$, $.79$ and $.35$ for the pooled correlation.

\paragraph{Leaving one system out.} The criteria file lists this check among the values reported without a prediction. Certify on $K-1$ systems and judge the regret on all $K$, the situation of a new policy (App.~\ref{app:newtasks}). Over every table, metric and left-out system, $10{,}599$ cases, the correlation's guarantee covers the regret in $10{,}462$ and the MMRV's in $10{,}451$; the realised regret on $K-1$ systems bounds the regret on $K$ in only $8{,}554$.

\paragraph{The MQM mapping.} The only modelling choice is the clipping at a non-translation on the $11$ MQM tables, and the findings do not rest on it. On the $21$ ESA tables alone, $417$ of $428$ correlations lie at or below $r_{\mathrm{vac}}(K)$,
the MMRV's guarantee is below the correlation's on $372$ of $428$ and never above, and the median
$\tau_b$ is $.41$ for the correlation against $.58$ for the MMRV. On the MQM tables these are $256$ of
$274$, $256$ of $274$, and $.26$ against $.47$.

\paragraph{Limitations of the audit.} A metric's cell is the mean of its segment scores, which is not
how every metric aggregates (BLEU is corpus-level). The outcome cells' sampling error is not modelled.
The metrics of a table are not independent, so the bootstrap is over tables, and coarse.

\subsection{Pretraining data: the DataDecide audit}\label{app:datadecide}

\paragraph{The decision.} A team often chooses its pretraining corpus by training small models on each
candidate and comparing them. DataDecide \citep{magnusson2025datadecide} releases $25$ corpora, each
used to train models of $14$ sizes from 4M to 1B parameters on three seeds, evaluated on the ten
multiple-choice benchmarks of OLMES \citep{gu2025olmes}. The policies are the corpora, the tasks the benchmarks, and the
outcome table holds the 1B models' accuracy. Each of the $13$ smaller sizes is an evaluator: its score
of a corpus on a benchmark is the accuracy of its model trained on that corpus.

\paragraph{Registration and construction.} \texttt{code/CRITERIA\_datadecide.md} fixes the decision, the
construction, two tolerances (one point, the primary one, and five points) and four hypotheses. It was
written before any accuracy was read. A cell is the primary accuracy at the last recorded step, averaged over seeds, with MMLU the mean of its
57 subtasks; the 1B outcomes are rounded to four places.

\paragraph{Results.} All four hypotheses held. $r_{\mathrm{vac}}(25)$ is $.941$ (rounded down), and $10$ of
the $13$ correlations lie at or below it. The correlation's instance bound is $.121$, the largest regret any choice can have, for every size, including the 530M models, whose correlation is
$.977$. The MMRV's instance bound is smaller for all $13$ sizes, from $.102$ at 4M to $.0501$ at 530M,
but no statistic certifies a per-benchmark choice within one point or five. Across sizes, $\tau_b$ with
minus the realised regret is $.79$ for the correlation, $.82$ for pairwise accuracy and $.87$ for the
MMRV. For deployment, the statistic DataDecide reports, decision accuracy (pairwise accuracy of the corpus
means), never certifies one point. Its instance bound is the whole spread of the corpus means, $.076$: ranking the worst corpus first misorders only $24$ of the $300$ pairs (accuracy $.92$), and no size exceeds $.89$. The realised deployment regret
is within one point for $4$ of the $13$ sizes.

\paragraph{Limitations.} The tasks are ten fixed benchmarks, not drawn at random, and the 1B accuracies are themselves
estimates on finite benchmarks, taken as true here. The last released checkpoint of the 750M models is at about $41\%$ of their token budget, for all three seeds; each size's three seeds are averaged at the same step.

\subsection{Reward models: best-of-32 on PPE}\label{app:ppe}

\paragraph{The decision.} Best-of-$K$ sampling returns, of $K$ answers to a question, the one a reward
model scores highest: routing, with items as tasks and the $32$ sampled answers as policies.
Preference Proxy Evaluations (PPE; \citealp{frick2025ppe}) releases, for five benchmarks with verifiable
answers (GPQA, IFEval, MATH, MBPP-Plus, MMLU-Pro), the correctness of each answer and the scores of $14$
reward models, a length baseline included. The sample number means nothing across questions, so
deployment and top-$k$ screening are not decisions here.

\paragraph{Registration and construction.} \texttt{code/CRITERIA\_ppe.md} fixes the construction, a
tolerance of five points and five hypotheses, before any score or label was read. The criteria did not foresee one case: a reward model prints a non-numeric score for $64$ answers, so their $31$ questions are
dropped for every evaluator. The tables have
$489$ to $510$ items: $70$ validations.

\paragraph{Results.} Four of the five hypotheses held. \emph{R-H2 failed:} the MMRV's instance bound
is below the correlation's on $1$ of the $70$ validations, not three quarters, because nearly every statistic guarantees only the largest possible regret, which is $1$ since PPE keeps only questions with both correct and incorrect answers; the exception is the MMRV of Athene-RM-70B on MMLU-Pro, at $.992$. Here $r_{\mathrm{vac}}(32)$ is $.954$ (rounded down) and the highest pooled correlation is $.473$. A reward model
does worse than always returning the same sample number, the best such rule in hindsight, on $30$ of
the $70$ validations. The median over benchmarks of $\tau_b$ with minus the realised regret is $.64$ for
the correlation, $.66$ for the MMRV and $.86$ for pairwise accuracy, a statistic PPE reports.

\subsection{Architectures: zero-cost proxies on NAS-Bench-201}\label{app:nas}

\paragraph{The decision.} Zero-cost proxies score candidate architectures without training, and only
the top-ranked ones are trained. NAS-Bench-Suite-Zero \citep{krishnakumar2022nasbenchsuitezero}
releases, for all $15{,}625$ architectures of NAS-Bench-201 \citep{dong2020nasbench201}, the validation accuracy on
CIFAR-10, CIFAR-100 and ImageNet16-120 and the scores of $13$ proxies. The policies are the
architectures, the tasks the datasets, and each proxy is an evaluator.

\paragraph{Registration and computation.} \texttt{code/CRITERIA\_nas.md} fixes the construction, two
tolerances (one and five points) and four hypotheses. With $15{,}625$ policies the exact programs run in double precision
(\texttt{sims/nas\_audit.py}), with per-option costs from sorted prefix sums. A guarantee is reported only if no choice's cost lies within a relative $10^{-9}$ of the budget; none does.

\paragraph{Results.} Three of the four hypotheses held. Here $r_{\mathrm{vac}}(15{,}625)$ is $.99990$
(rounded down) and the Spearman threshold of Theorem~\ref{thm:rankpooled}, $1-r_{\mathrm S}\ge4/K^3$, is met by every proxy, so neither pooled coefficient guarantees anything from its value;
the highest pooled Pearson and Spearman coefficients are $.607$ and $.942$. With the table, every proxy's correlation
guarantees only the largest per-dataset regret, $.670$, and for deployment the whole spread of the
architecture means, $.665$. The MMRV's guarantee is below the correlation's for all $13$ proxies, but
no lower than $.335$. Realised regrets are small: $.020$ for the best proxy, $.054$ for the median. \emph{N-H4 failed:} Precision@100, which NAS-Bench-Suite-Zero also reports, orders the proxies by the regret of training their top $100$ better than Spearman's coefficient, with $\tau_b$ $.79$ against $.22$: the statistic that scores the decision predicts
its regret, as the MMRV and pairwise accuracy do in the other audits (here the MMRV does worse than
$r$, Sec.~\ref{sec:exchange}).

%% file: app_trials.tex
\section{Trial counts, new tasks and policies, and testing a top-\texorpdfstring{$k$}{k} screen}\label{app:trials}

\subsection{The least regret that can be guaranteed when the outcome table is uncertain}\label{app:hinge}

How far may the published rates be off before no choice, even a perfect evaluator's, can be
guaranteed? Let each true rate lie within a \emph{radius} $\eta$ of the published one: put
$u^{-}_{it}=\max(0,p_{it}-\eta)$ and $u^{+}_{it}=\min(1,p_{it}+\eta)$, and let the \emph{box}
$\mathcal U_\eta$ be the set of tables lying entrywise between them. The radius is an assumption stated about the outcome table, not an estimate.

A choice $a$ is \emph{guaranteed
over the box} at tolerance $\varepsilon$ when its regret is at most $\varepsilon$ for every table in
$\mathcal U_\eta$. The worst table sets, task by task, every other policy to $u^{+}$ and the chosen one
to $u^{-}$, so the largest regret the box assigns to $a$ is
$W_{\mathcal U_\eta}(a)=\frac1T\sum_t\omega_{a(t)t}$, with
$\omega_{it}=\max(0,\max_{j\neq i}u^{+}_{jt}-u^{-}_{it})$. The best choice can also be made task by
task, so the \emph{hinge}, the minimax regret over the box, is
\begin{equation}\label{eq:hinge}
M_{\mathcal U_\eta}
=\frac1T\sum_{t=1}^{T}\Bigl(\min_i\ \max_{j\ne i}\bigl(u^{+}_{jt}-u^{-}_{it}\bigr)\Bigr)_{\!+}
=\frac1T\sum_{t=1}^{T}\min_i\omega_{it},
\end{equation}
with $(x)_+=\max(0,x)$. No choice can be guaranteed at tolerance $\varepsilon$ once
$M_{\mathcal U_\eta}>\varepsilon$.

\begin{lemma}[the hinge of an uncertain outcome table]\label{lem:hinge}
The minimum in \eqref{eq:hinge} is attained at the outcome table's own leader. If
$p_{(2)t}+\eta\le1$ and $p_{(1)t}\ge\eta$ for every task, then $M_{\mathcal U_\eta}=\frac1T\sum_{t}(2\eta-g_t)_{+}$.
The map $\eta\mapsto M_{\mathcal U_\eta}$ is nondecreasing and piecewise linear, with
breakpoints only at $g_t/2$, $1-p_{(2)t}$ and $p_{(1)t}$. Hence the largest radius
$\eta^\star(\varepsilon)$ at which some choice can be guaranteed at a tolerance $\varepsilon<1$ is attained, and is rational when every entry of the outcome table is.
\end{lemma}

\begin{proof}[Proof of Lemma~\ref{lem:hinge}]
Let $i^{\star}$ be a leader of column $t$ and let $i\neq i^{\star}$; then $\{j\neq i\}$ contains
$i^{\star}$, so
$\max_{j\neq i}u^{+}_{jt}=\max_ju^{+}_{jt}\ge\max_{j\neq i^{\star}}u^{+}_{jt}$,
while $u^{-}_{it}\le u^{-}_{i^{\star}t}$ because truncation to $[0,1]$ is nondecreasing, whence
$\omega_{it}\ge \omega_{i^{\star}t}$: any other policy faces a stronger rival from a lower floor.
Since the largest entry other than a leader is $p_{(2)t}$,
$\min_i\omega_{it}=\max(0,\min(1,p_{(2)t}+\eta)-\max(0,p_{(1)t}-\eta))$, which is $(2\eta-g_t)_+$ when
neither truncation is active, that is, under the stated hypothesis.
As a function of $\eta$, each $\min_i\omega_{it}$ is the positive part of a nondecreasing piecewise
linear function with breakpoints at $\eta=1-p_{(2)t}$ and $\eta=p_{(1)t}$, and its zero crossing is at
$\eta=g_t/2$, which precedes both since $g_t\le p_{(1)t}$ and $g_t\le1-p_{(2)t}$. So the breakpoints of
$M_{\mathcal U_\eta}$ are among the three listed values, and it rises continuously from
$M_{\mathcal U_0}=0$ to $M_{\mathcal U_1}=1$. Some choice is guaranteed at tolerance $\varepsilon$
exactly when $M_{\mathcal U_\eta}\le\varepsilon$, so the largest such $\eta$ is attained; it is found by
locating the two consecutive breakpoints on either side and solving one affine equation, which returns a
rational number whenever the outcome table is rational.
\end{proof}

 Across the nine outcome tables, the largest radius at which some choice
is guaranteed within five points ranges from $.0299$ (WMT23) to $.2028$ (WorldGym); a brute-force check
on $32$ small tables confirms the lemma.

\paragraph{Resampling at the published trial counts.} Five of the eight source tables
(App.~\ref{app:corpus}) state a per-cell trial count, from ten on WorldGym to seventy-five on a SIMPLER
task; they come from four distinct real experiments. Take the published rates as true and resample every cell's trials. The probability that choosing each task's observed leader loses at most five points,
computed exactly, is $.969$ on WorldGym, $.909$ on SIMPLER Google Robot, $.928$ on WidowX and $.997$ on AutoEval; only AutoEval, with fifty trials per cell, reaches $.99$.
\label{app:evidence}A perfect evaluator, scoring the published rates, has median correlation $.947$,
$.987$, $.947$ and $.995$ with $200$ resampled tables on WorldGym, SIMPLER Google Robot, WidowX and
AutoEval. It exceeds each resampled table's own five-point required correlation $r_{.05}$ (Eq.~\eqref{eq:exchange} gives at most five points exactly when $r>r_{.05}$) on $0$, $0$, $30$ and $31$ of them, while
its MMRV guarantees five points on $198$, $190$, $191$ and $199$.

\paragraph{The hinge at a stated confidence.} Now the box is every outcome table whose cells lie in their
exact (Clopper--Pearson) binomial intervals at their trial counts, WorldGym's half-credit totals rounded to a count; a Wald interval would give radius zero to the
thirty-two cells with no successes or no failures. The level is per cell; a simultaneous region would
be wider and could only remove guaranteed choices. At $95\%$ the hinges are $.2090$ on WorldGym ($.192$ to $.242$ over its roundings),
$.1445$ on SIMPLER's Google Robot, $.2166$ on WidowX and $.0798$ on AutoEval, all above $.05$: on none
of the four designs can any choice be guaranteed within five points, even by a perfect evaluator. The
four-task subset of AutoEval used for its SIMPLER comparison has a hinge of $.0481$, but reuses AutoEval's tasks.

\subsection{The evaluator's own rollouts: a simultaneous confidence box}\label{app:joint}

The box above fixes the evaluator's scores and lets only the outcomes vary, so it bounds the regret
of the choice the published scores made. A simulator is validated to be trusted on later runs, whose rollouts will differ. The claim of interest is therefore the choice the evaluator makes when run to convergence: the leader of its true table $Y$, which the published scores estimate. When both tables are estimated from trials, both must be boxed.

Let every cell of the outcome table $P$ and of $Y$ come with an interval,
$[P^-_{it},P^+_{it}]$ and $[Y^-_{it},Y^+_{it}]$, that covers the cell's true rate with probability at
least $1-\alpha/(2KT)$. Let $a^\star$ choose on each task the leader of $Y$, and $s^\star$ deploy the
policy with the largest mean of $Y$ over tasks, ties broken against the evaluator in both. Put, with $\Gamma_t$ the \emph{contenders} for $Y$'s leader on task $t$,
\begin{gather*}
\Gamma_t=\bigl\{i:\ Y^+_{it}\ge\textstyle\max_{j\ne i}Y^-_{jt}\bigr\},\qquad
\omega_{it}=\bigl(\textstyle\max_{j\ne i}P^+_{jt}-P^-_{it}\bigr)_+,\qquad
\lambda_{it}=\bigl(\textstyle\max_{j\ne i}P^-_{jt}-P^+_{it}\bigr)_+,\\
R^+=\frac1T\sum_{t}\max_{i\in \Gamma_t}\omega_{it},\qquad
R^-=\frac1T\sum_{t}\min_{i\in \Gamma_t}\lambda_{it},
\end{gather*}
and define $\Gamma$, $R^+_{\mathrm{dep}}$ and $R^-_{\mathrm{dep}}$ in the same way from the means over tasks
of the four bounds, $\bar Y^\pm_i=\frac1T\sum_tY^\pm_{it}$ and $\bar P^\pm_i=\frac1T\sum_tP^\pm_{it}$.

\begin{theorem}[a simultaneous confidence box]\label{thm:joint}
With probability at least $1-\alpha$, simultaneously
$R^-\le R(a^\star)\le R^+$ and $R^-_{\mathrm{dep}}\le R_{\mathrm{dep}}(s^\star)\le R^+_{\mathrm{dep}}$,
where both regrets are taken on the true outcome table. This needs no independence between cells.
Each upper end is attained by some pair of tables in the box, so no bound computed from the box alone
is smaller.
\end{theorem}

\begin{proof}
By the union bound over the $2KT$ cells, the event $E$ that every interval covers its true rate has
probability at least $1-\alpha$. On $E$, the evaluator's choice
$i=a^\star(t)$ satisfies $Y^+_{it}\ge Y_{it}\ge\max_{j\ne i}Y_{jt}\ge\max_{j\ne i}Y^-_{jt}$, so
$a^\star(t)\in \Gamma_t$. The regret of policy $i$ on task $t$ is $(\max_{j\ne i}P_{jt}-P_{it})_+$, which is
nondecreasing in every $P_{jt}$ with $j\ne i$ and nonincreasing in $P_{it}$, so on $E$ it lies in
$[\lambda_{it},\omega_{it}]$. Taking the worst and the best contender on each task and averaging gives
$R^-\le R(a^\star)\le R^+$. The policy means of the two true tables lie between the means of the bounds,
and the same argument on the means gives the deployment statement.

For attainment, fix on each task a contender $i$ that maximises $\omega_{it}$ over $\Gamma_t$. Put
$Y_{it}=Y^+_{it}$ and $Y_{jt}=Y^-_{jt}$ for $j\ne i$, so that $i$ leads $Y$, possibly tied, and put
$P_{it}=P^-_{it}$ and $P_{jt}=P^+_{jt}$ for $j\ne i$. If $i$ is chosen, its regret is $\omega_{it}$.
A tied $j$ has $Y^+_{jt}\ge Y^-_{jt}=Y^+_{it}\ge Y^-_{kt}$ for every $k\ne j$, so $j\in \Gamma_t$; if it is chosen, its regret is at least $\omega_{it}$ by the tie rule and at most
$\omega_{jt}\le\omega_{it}$, since the box's tables have $P_{kt}\le P^+_{kt}$ and $P_{jt}\ge P^-_{jt}$.
Hence the regret is $\max_{i\in \Gamma_t}\omega_{it}$ on every task, and its mean is $R^+$. Deployment is the
same construction on whole rows.
\end{proof}

Every interval here is exact: for a cell scored in successes, the Clopper--Pearson interval. On
WorldGym a trial scores $0$, $\frac12$ or $1$ on both sides; writing the score as $(A+B)/2$, with
$A$ the indicator of at least half credit and $B$ of full credit, the cell's interval is the hull, over
every split of the published total between $A$ and $B$, of the mean of the two exact intervals, each with half the cell's miss probability. By the union bound it covers the true rate whatever the split.

The theorem needs the trial counts of both tables. Three validations state them: WorldGym, with ten real
trials and one world-model rollout per real trial; AutoEval's autonomous scorer; and SIMPLER on
AutoEval's four tasks, both of which print counts out of fifty. SIMPLER's own source and Ctrl-World
do not state their simulated rollouts per cell. At $\alpha=.05$, simultaneous over all cells:
\begin{itemize}[leftmargin=*,topsep=1pt,itemsep=1pt,parsep=0pt]
\item \emph{WorldGym.} Every policy remains a contender on every task, and $R^+=.971$.
\item \emph{AutoEval's autonomous scorer.} Its published routing choices have no regret, yet the box
guarantees only $R^+=.297$ for its own choice, and $R^+_{\mathrm{dep}}=.391$ for deployment.
\item \emph{SIMPLER on AutoEval's tasks.} SIMPLER is certified to lose: $R^-=.041$. On \emph{To Basket}
its only contender is Open~$\pi_0$, which loses at least $16.5$ points to MiniVLA.
\end{itemize}
At per-cell $95\%$, without the Bonferroni correction, the per-task bounds become $.913$, $.232$ and $R^-=.210$.

Suppose both tables showed their published rates at $m=100,200,\dots,3200$ trials per cell. $R^+$ stays within five points from $m=1000$ on for the autonomous scorer ($R^+=.0499$), twenty times its
fifty trials, and from $m=400$ for WorldGym, forty times its ten. For SIMPLER it never is: its certified
loss grows to $R^-=.314$ at $m=3200$. At rates like these, certifying the evaluator rather than one draw
of its scores takes hundreds to thousands of rollouts and real trials per cell
(\texttt{sims/joint\_box.py}).

\subsection{New tasks and new policies}\label{app:newtasks}

A guarantee on the published table transfers to new tasks drawn at random, at a cost in the number
of tasks; for exchangeable policies, no guarantee covers a new one unless there are many.

\begin{theorem}[a guarantee on new tasks]\label{thm:newtasks}
Let the $T$ published tasks be independent draws from a law $\mathcal D$, each an outcome column
and a score column over the same $K$ policies. On a task, let $\ell\in[0,1]$ be the loss of the
policy the evaluator scores highest, and $R(\mathcal D)=\E_{\mathcal D}\ell$. Let $\widehat B$ be any
function of the published tables with $\frac1T\sum_t\ell_t\le\widehat B$ on every outcome table, such
as the MMRV's instance bound or $\varepsilon^\star(r)$, and $c=\log(1/\alpha)/(2T)$. With probability at
least $1-\alpha$, $R(\mathcal D)\le\bigl(\sqrt{\widehat B+c}+\sqrt c\bigr)^2$.
\end{theorem}

\begin{proof}
The losses are independent and lie in $[0,1]$, so $\E\ell^2\le\E\ell=R$, and the lower-tail
inequality for nonnegative variables \citep{maurer2006concentration} gives
$\Pr[\frac1T\sum_t\ell_t\le R-x]\le\exp(-Tx^2/(2R))$. Taking $x=2\sqrt{cR}$ makes the right-hand side
$\alpha$, so with probability at least $1-\alpha$, $R\le\frac1T\sum_t\ell_t+2\sqrt{cR}\le\widehat
B+2\sqrt{cR}$, a quadratic inequality in $\sqrt R$ whose solution is $\sqrt R\le\sqrt c+\sqrt{\widehat
B+c}$.
\end{proof}

\begin{corollary}[how many tasks a transfer needs]\label{cor:tasks}
For $0\le\widehat B<\varepsilon$, the bound of Theorem~\ref{thm:newtasks} is at most $\varepsilon$
exactly when $T\ge2\varepsilon\log(1/\alpha)/(\varepsilon-\widehat B)^2$. At $\varepsilon=.05$ and
$\alpha=.05$ this is $120$ tasks when $\widehat B=0$, $188$ when $\widehat B=.01$ and $333$ when
$\widehat B=.02$.
\end{corollary}

\begin{proof}
$(\sqrt{\widehat B+c}+\sqrt c)^2\le\varepsilon$ if and only if $\sqrt{\widehat B+c}\le\sqrt\varepsilon-\sqrt
c$, that is $\widehat B\le\varepsilon-2\sqrt{\varepsilon c}$, that is $c\le(\varepsilon-\widehat
B)^2/(4\varepsilon)$; substitute $c=\log(1/\alpha)/(2T)$.
\end{proof}

At the published task counts, four to $17$, with each validation's MMRV instance bound as $\widehat B$ and $\alpha=.05$, the bound is at least $.48$ on every corpus validation, so it certifies nothing there; it serves as a design rule for random tasks, and a sharper bound could need fewer.

\paragraph{A new policy.} Now consider policies instead of tasks: $K+1$ exchangeable rows (their joint law is unchanged
by reordering, as when policies are drawn from a common population), of which $K$ are published. Write $z_i=(p_{i\cdot},y_{i\cdot})$ for policy $i$'s row of outcomes in $[0,1]^T$ and scores
in $\mathbb R^T$. The published rows come in a uniformly random order. Let $R_{K+1}$ be
the per-task regret of the taskwise maximiser of the scores over all $K+1$ rows, ties resolved by any
fixed rule. A guarantee $\widehat\varepsilon$, a function of the published rows and of independent
randomness, is valid at level $\alpha$ if $\Pr[R_{K+1}\le\widehat\varepsilon]\ge1-\alpha$ under every
exchangeable law of $(z_1,\dots,z_{K+1})$.

\begin{theorem}[a new policy]\label{thm:newpolicy}
Let $\widehat\varepsilon$ be valid at level $\alpha$. Then on every published table $(p,y)$, $\Pr[\widehat\varepsilon<R_{\max}(p)]\le(K+1)\alpha$, where $R_{\max}(p)=\frac1T\sum_t\max_ip_{it}$.
\end{theorem}

Since $R_{\max}(p)$ is at least the largest regret among the published policies, a guarantee that reaches
it rules out no loss among them; an always nontrivial $95\%$ claim thus needs at least $19$ validated
policies ($9$ at $90\%$). The same exchangeability makes a split-conformal interval from $K$ calibration points infinite when $K<1/\alpha-1$ \citep{vovk2005algorithmic}.

\begin{proof}[Proof of Theorem~\ref{thm:newpolicy}]
Fix a published outcome table $p$ with scores $y$ and rows $c_1,\dots,c_K$. Let $w$ be the row with outcome
$0$ on every task and score $1+\max_{i,t}y_{it}$ on every task, a decoy that always fails but is scored
above everything. Let the $K+1$ rows be
$c_1,\dots,c_K,w$ in a uniformly random order, an exchangeable law. Whatever the order, the scores
put $w$ strictly first on every task, so the choice takes $w$ everywhere and
$R_{K+1}=\frac1T\sum_t(\max\{\max_ip_{it},0\}-0)=R_{\max}(p)$. With probability $1/(K+1)$ the unpublished
row is $w$. Then the published rows are $c_1,\dots,c_K$ in a uniformly random order, so
$\widehat\varepsilon$ has the same law as on $(p,y)$, and the guarantee fails whenever
$\widehat\varepsilon<R_{\max}(p)$. Since $\widehat\varepsilon$ is valid at level $\alpha$,
$\alpha\ge\Pr[R_{K+1}>\widehat\varepsilon]\ge\Pr[\widehat\varepsilon(p,y)<R_{\max}(p)]/(K+1)$.
\end{proof}

The bound holds for every $T$ and every outcome table, even one the evaluator ranks perfectly. It is a worst case, reached through a decoy: on WMT, with up to twenty systems, guarantees computed without one system covered the regret with it in about $99\%$ of cases (App.~\ref{app:wmtaudit}).

\subsection{Testing a screen}\label{app:practice}

Testing a screen on hardware is a ranking-and-selection problem \citep{bechhofer1954,kim2006selecting}. We use its simplest single-stage procedure, equal trials per policy, and guarantee a good selection (within $\varepsilon$ of the best) rather than a correct one under an indifference zone; in best-arm identification this is $(\varepsilon,\delta)$-PAC.

\paragraph{Setting.} Here $p$ denotes the true success rates and $\hat p$ an estimate of them built from
trials: the published table, or the table a hardware test produces. Regret is measured against the
policy with the highest mean. Let $s'$ be the chosen policy, or the screened-in policy with the largest $\hat p_i$. A \emph{published guarantee} $\widehat B$ is any function of $\hat p$ and the scores with $\max_i\hat p_i-\hat p_{s'}\le\widehat B$. The correlation's screening guarantee (App.~\ref{app:coarse}) is one; $\widehat B=0$ describes a hardware test with no evaluator. Trial $j$ of policy $i$ on task $t$ has
an outcome $X_{itj}\in[0,1]$ with mean $p_{it}$, trials are independent given the tasks, policy $i$
receives $n_t$ trials on task $t$, fixed in advance, and $\hat p_i=\frac1T\sum_t\frac1{n_t}\sum_jX_{itj}$.
The \emph{effective number of trials} per policy is $\nu=T^2/\sum_tn_t^{-1}$, which is $Tn$ when every
task receives $n$. Every bound below is sufficient, not necessary.

\begin{lemma}[resolving the published policies]\label{lem:resolve}
Let $\mu_i=\frac1T\sum_tp_{it}$ and let $b$ be a policy with the largest $\mu_i$. With probability at
least $1-(K-1)e^{-\nu x^2}$, every choice or top-$k$ screen with published guarantee $\widehat B$ is within
$\widehat B+x$ of $\mu_b$.
\end{lemma}

\begin{proof}
Fix $s\ne b$. The difference $\hat p_b-\hat p_s$ is a sum of independent terms, one per trial; a trial
on task $t$ contributes a term of range $1/(Tn_t)$, so the squared ranges sum to
$2\sum_tn_t/(Tn_t)^2=2/\nu$. Hoeffding's inequality \citep{hoeffding1963} gives
$\Pr[(\mu_b-\mu_s)-(\hat p_b-\hat p_s)\ge x]\le e^{-2x^2/(2/\nu)}=e^{-\nu x^2}$, and a union bound over
the $K-1$ policies $s\ne b$ makes all these differences at most $x$ at once. On that event,
$\mu_b-\mu_{s'}\le\hat p_b-\hat p_{s'}+x\le\max_i\hat p_i-\hat p_{s'}+x\le\widehat B+x$.
\end{proof}

\paragraph{The screening protocol.}
Fix a tolerance $\varepsilon=\varepsilon_1+\varepsilon_2$ and an error probability $\delta$:
\begin{enumerate}[leftmargin=*,topsep=1pt,itemsep=1pt,parsep=0pt]
\item Take the evaluator's top $k$ policies, with $k$ the smallest size whose screening guarantee on the published tables is at most $\varepsilon_1$ (App.~\ref{app:coarse}).
\item Test each with $\lceil\log((k-1)/\delta)/\varepsilon_2^2\rceil$ real trials, rounded up to a multiple of $T$ and spread evenly over the tasks, so that each observed rate estimates the mean over tasks.
\item Deploy the best observed one. Taking the published rates as true for step 1, it is within $\varepsilon$ of the best policy with probability at least $1-\delta$ (Lemma~\ref{lem:resolve} with $k$ for $K$ and $\widehat B=0$).
\end{enumerate}

\paragraph{What the evaluator saves.} We compare two ways to deploy a policy within five points of the best with probability $.95$: test all $K$ policies, or test the evaluator's top $k$, the five points split between screen and test at the cheapest split. For a fixed $k$ the test never gets cheaper as the screen takes more of the tolerance, so the cheapest split gives the screen exactly the guarantee of some $k$ (screening down to one policy deploys it untested). These figures take the published rates as exact at screening, which they are not (App.~\ref{app:hinge}). With that reservation, Table~\ref{tab:decisions} gives the share of hardware trials saved. A benchmark should therefore also report $k^\star$ at a tolerance finer than the laboratory's.